\documentclass[fleqn,10pt,sort]{wlscirep}
\usepackage[utf8]{inputenc} 
\usepackage[T1]{fontenc}    
\usepackage{hyperref}       
\usepackage{url}            
\usepackage{booktabs}       
\usepackage{amsfonts}       
\usepackage{nicefrac}       
\usepackage{microtype}      
\usepackage{lipsum}
\usepackage{fancyhdr}       
\usepackage{graphicx}       
\usepackage{amssymb, amsmath, bm, amsfonts}
\usepackage{float,graphbox}
\usepackage{epstopdf}
\usepackage{subcaption}
\usepackage{hyperref}
\usepackage{color, xcolor}
\usepackage{booktabs, multirow, makecell}
\usepackage{threeparttable}
\usepackage{array,tabularray}
\usepackage{titlesec}
\usepackage[ruled,linesnumbered]{algorithm2e}
\usepackage{enumitem}
\usepackage{tikz}
\setlist{nosep}
\usepackage{soul}
\usepackage{lipsum}

\usepackage{amsthm}

\usepackage{newtxmath}

\usepackage[sectionbib]{bibunits} 
\defaultbibliographystyle{naturemag-doi}

\fancypagestyle{maintext}{
    \lhead{}%
    \chead{}%
    \rhead{}%
    \lfoot{}%
    \cfoot{}%
    \rfoot{\small\sffamily\bfseries\thepage/\pageref{MainTextEnd}}%
}

\fancypagestyle{SI}{
    \lhead{}%
    \chead{}%
    \rhead{}%
    \lfoot{}%
    \cfoot{}%
    \rfoot{\small\sffamily\bfseries\thepage/\pageref{SIEnd}}%
}

\usepackage[figuresleft]{rotating}
\graphicspath{{figures/}}
\title{Enforcing hidden physics in physics-informed neural networks}

\author[1]{Nanxi Chen}
\author[2]{Sifan Wang}
\author[1,*]{Rujin Ma}
\author[1]{Airong Chen}
\author[3,*]{Chuanjie Cui}
\affil[1]{Tongji University, College of Civil Engineering, Shanghai, 200092, China}
\affil[2]{Yale University, Institute for Foundations of Data Science, New Haven, CT 06520, USA}
\affil[3]{University of Oxford, Department of Engineering Science, Oxford, OX1 3PJ, UK}

\affil[*]{Corresponding authors. rjma@tongji.edu.cn; chuanjie.cui@eng.ox.ac.uk}

\begin{abstract}
Physics-informed neural networks (PINNs) represent a new paradigm for solving partial differential equations (PDEs) by integrating physical laws into the learning process of neural networks. However, ensuring that such frameworks fully reflect the physical structure embedded in the governing equations remains an open challenge, particularly for maintaining robustness across diverse scientific problems.
In this work, we address this issue by introducing a simple, generalized, yet robust irreversibility-regularized strategy that enforces hidden physical laws as soft constraints during training, thereby recovering the missing physics associated with irreversible processes in the conventional PINN. This approach ensures that the learned solutions consistently respect the intrinsic one-way nature of irreversible physical processes. Across a wide range of benchmarks spanning traveling wave propagation, steady combustion, ice melting, corrosion evolution, and crack growth, we observe substantial performance improvements over the conventional PINN, demonstrating that our regularization scheme reduces predictive errors by more than an order of magnitude, while requiring only minimal modification to existing PINN frameworks. 
\end{abstract}
\begin{document}

\raggedbottom
\maketitle
%
%
\pagestyle{maintext}
\thispagestyle{empty}

\begin{bibunit}[naturemag-doi] 

\section*{Introduction}

Physics-informed neural networks (PINNs) have emerged as a powerful paradigm in scientific machine learning by embedding physical laws into neural network training through minimizing physics-based loss functions \cite{raissiPhysicsinformedNeuralNetworks2019}. These losses act as soft constraints, guiding the model to produce solutions that satisfy the governing equations, with or without experimental or simulated data. Owing to their elegance, feasibility, and versatility, PINNs have been widely recognized as a promising framework for solving both forward and inverse problems governed by partial differential equations (PDEs), with successful demonstrations across fluid mechanics \cite{raissi2020hidden,almajid2022prediction,eivazi2022physics,cao2024surrogate, wang2025simulating}, heat transfer \cite{xu2023physics,bararnia2022application,gokhale2022physics}, bioengineering \cite{kissas2020machine,zhang2023physics,caforio2024physics}, materials science \cite{zhang2022analyses,jeong2023physics,hu2024physics,chenPFPINNsPhysicsinformedNeural2025,chenSharpPINNsStaggeredHardconstrained2025}, electromagnetics \cite{kovacs2022conditional,khan2022physics,baldan2023physics}, and geosciences \cite{smith2022hyposvi,song2023simulating,ren2024seismicnet,wang2025deep}. 
Although PINNs have achieved notable success across a range of applications, they frequently suffer from slow convergence and limited accuracy \cite{luo2025physics,zhang2026physics}. These shortcomings become especially pronounced in dealing with complex systems, thereby limiting their reliability as forward solvers for PDEs.

This has motivated a growing body of work aimed at addressing the underlying challenge, with advances emerging along several fronts. For example, on the architectural side, innovations include novel network backbones \cite{wang2021understanding,sitzmann2020implicit,fathony2021multiplicative,moseley2021finite,kang2022pixel,cho2024separable,wang2024piratenets}, adaptive activation functions \cite{jagtap2020adaptive,abbasi2024physical}, and expressive coordinate embeddings \cite{wang2021eigenvector,costabal2024delta,zeng2024rbf,huang2024efficient}. In parallel, training-related improvements include adaptive sampling of collocation points \cite{nabian2021efficient,daw2022rethinking,wu2023comprehensive}, advanced optimization algorithms \cite{muller2023achieving,jnini2024gauss,song2024admm,urban2025unveiling, kiyani2025optimizing, wang2025gradient}, coupled-automatic–numerical differentiation framework \cite{chiu2022can}, gradient-enhanced PINNs \cite{yu2022gradient}, and progressive training schemes such as sequential \cite{wight2020solving,krishnapriyan2021characterizing,cao2023tsonn} and transfer learning \cite{desai2021one,goswami2020transfer,chakraborty2021transfer}. However, these efforts mainly focus on the machine learning and optimization aspects, with relatively little attention given to strengthening physical consistency—an element that is fundamental to accurate PDE modeling and central to the success of classical methods such as the finite element method (FEM).

\begin{figure}[htbp]
    \centering
    \includegraphics[width=\textwidth]{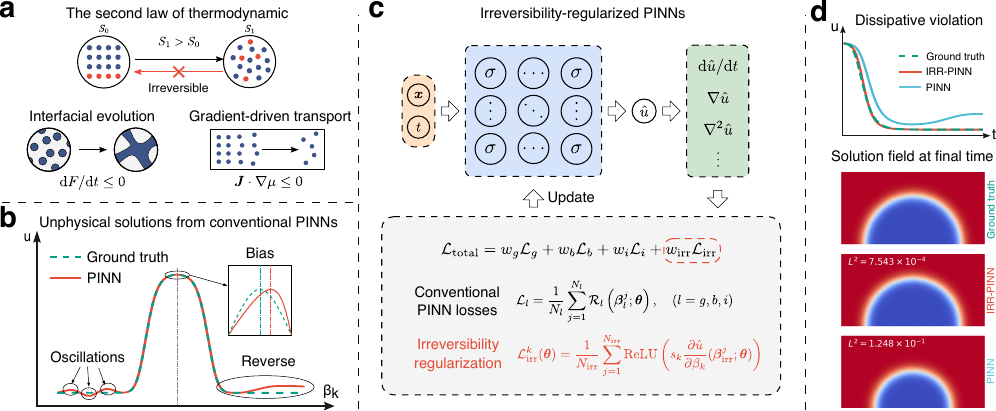}
    \caption{
        \textbf{Overview of the irreversibility-regularized PINNs}. 
        (\textbf{a}) The second law of thermodynamics governs many physical processes characterized by inherent irreversibility, such as interfacial evolution and gradient-driven transport. However, this irreversible behavior is typically \emph{implicitly} encoded in the governing PDEs. 
        (\textbf{b}) Conventional PINNs, which only minimize the PDE residuals and initial/boundary conditions without respecting this hidden irreversibility, may produce unphysical solutions such as localized oscillations, biased extrema, or reversed fronts, violating the intrinsic irreversible nature.
        (\textbf{c}) We introduce a simple, generalized, yet robust irreversibility-regularized loss term to explicitly enforce the hidden physical irreversibility as a soft constraint during training. The point-wise violation measure $V_k=\text{ReLU}\left(s_k \partial u / \partial \beta_k\right)$ penalizes any local violations of the irreversibility condition, ($s_k \in \{+1, -1\}$ indicating forward or backward irreversibility along the coordinate $\beta_k$). This regularization term is seamlessly integrated into the conventional PINN loss function, guiding the neural network solution toward physically consistent results.
        (\textbf{d}) The irreversibility regularization effectively suppresses unphysical violations, yielding substantial improvements in both accuracy and physical consistency in a benchmark test.
    }
    \label{fig:framework}
\end{figure}

From a physical perspective, this difficulty arises from the fundamentally different paradigm that PINNs follow compared with classical numerical solvers. Traditional FEM discretize the governing equations into algebraic systems with well-established accuracy and convergence guarantees, whereas PINNs reformulate the solving process as an optimization task, training neural networks to minimize the PDE residuals. This provides considerable flexibility, but offers far less rigorous error control than classical numerical methods. Put more simply, as shown in Figures \ref{fig:framework}a and \ref{fig:framework}b, PINNs can learn only the \emph{explicit} components of a PDE, that is, those that appear directly in the loss function. \emph{Implicit} physical constraints, such as the irreversible behaviors required by \emph{the Second Law of Thermodynamics}, cannot be automatically preserved as they are not encoded in the loss function. For example, in the point-source diffusion problem governed by \emph{the Second Fick's Law}, the PDE and initial and boundary conditions describe how an initially concentrated pulse of material spreads outward uniformly in all directions and progressively smooths over time. The true physical solution is therefore a radially decreasing Gaussian profile. Any deviation from this monotonic decay implies a reversal of the concentration gradient and thus an unphysical flux from low to high concentration, leading to negative local entropy production and violating \emph{the Second Law of Thermodynamics}. However, since this constraint is not explicitly represented in the governing PDE, the network may develop small non-physical oscillations and/or reversed fronts (Figure \ref{fig:framework}b), thereby degrading predictive accuracy, increasing training cost, and even leading to training failure.

One popular way to address the hidden physics is to reformulate the problem within an energy-minimization framework, such as Deep Energy Method \cite{samaniego2020energy}, Deep Ritz Method \cite{yu2018deep}, and Thermodynamically-consistent PINNs \cite{patel2022thermodynamically,kutukEnergyDissipationPreserving2025}. These approaches train the neural network by minimizing a physically meaningful energy functional rather than strong-form PDE residuals, thereby inherently enforcing the energy-dissipation property, which is consistent with \emph{the Second Law of Thermodynamics}. However, the applicability of energy-based formulations is limited to physical systems for which a thermodynamic potential or a minimum-energy principle can be clearly defined. Consequently, such approaches cannot be directly extended to non-potential systems such as fluid mechanics \cite{he2023deep}. In addition, although possible through penalty or trial-function strategies, enforcing Dirichlet boundary conditions in energy-minimization frameworks is not straightforward and often requires problem-specific treatment \cite{muller2019deep}. In some special cases, these physically-enhanced methods are computationally expensive to train, sometimes much less efficient than conventional finite element methods \cite{he2023deep}.

In this study, we first propose a simple yet arguably more applicable strategy to incorporate the hidden physical laws into PINNs. Rather than constructing an explicit thermodynamic potential, our approach directly target the physical representation of these hidden laws, namely, irreversibility. Specifically, many natural processes show an intrinsic directionality that cannot spontaneously reverse without external intervention. Motivated by this universal nature, we introduce a practical regularization technique to bridge the gap between the hidden physical irreversibility and conventional strong-form PINNs (Figure \ref{fig:framework}), and we examine its impact on optimization dynamics, accuracy, and convergence. Through extensive benchmark evaluations, we demonstrate that this regularization strategy can reduce predictive errors by more than an order of magnitude without any sacrifice in computational cost. Importantly, our approach is neither task-specific nor model-specific, and can be readily applied to various scientific areas and physics-informed learning frameworks, opening a new pathway for trustworthy scientific machine learning. 

\section*{Results}

We aim to outline the core concept and demonstrate the performance of our proposed regularization strategy, which enforces hidden irreversibility within PINNs through a simple yet robust formulation. We evaluate its effectiveness across five benchmark problems that span a broad range of physical systems, including traveling-wave propagation, steady combustion, ice melting, corrosion modeling, and crack growth. Together, these benchmarks cover both directional and dissipative forms of irreversibility and allow us to assess the accuracy, stability, and physical consistency achieved by the IRR-PINN framework.

\subsection*{Irreversibility-regularized PINNs}

We develop a regularization strategy that enforces the hidden irreversibility as a soft constraint within the conventional PINN, with its flowchart shown in Figure \ref{fig:framework}. The strategy is simple and broadly applicable, and it remains fully consistent with the PINN paradigm, where the governing equations, initial and boundary conditions, and other physical principles are encoded as loss terms to be jointly minimized. Formally, the total loss $\mathcal{L}_{\text{total}}$ can be expressed as  
\begin{equation}
    \mathcal{L}_{\text{total}}
    = w_g \mathcal{L}_g + w_b \mathcal{L}_b + w_i \mathcal{L}_i
    + \boxed{w_{\text{irr}} \mathcal{L}_{\text{irr}}}\,\,,
\label{eq:augmented_loss}
\end{equation}
where $w_g$, $w_b$, $w_i$, and $\mathcal{L}_g$, $\mathcal{L}_b$, $\mathcal{L}_i$ are weights and loss terms associated with the PDE residuals and the boundary and initial conditions, identical to those used in the conventional PINN (see Section~\ref{sec:basic-pinns-formulation} in the Supplementary Information). The last term $w_{\text{irr}} \mathcal{L}_{\text{irr}}$ guides the neural network solution toward satisfying the irreversibility constraints, thereby ensuring physically consistent results that respect the inherent directionality of the underlying physical processes. 

To formulate the loss term $\mathcal{L}_{\text{irr}}$, we consider a computational domain $\mathcal{D} = \Omega \times [0,T] \times \mathcal{P}$, where $\Omega \subset \mathbb{R}^d$, $[0,T]$, and $\mathcal{P}$ represent the spatial domain, temporal domain, and parameter space, respectively. Let $u\colon \mathcal{D} \to \mathbb{R}$ be a physical field defined over the generalized coordinates $\bm{\beta} = (\beta_1, \beta_2, \ldots, \beta_n) \in \mathcal{D}$, which may include spatial coordinates, time, and other parameters governing the system dynamics. With these definitions, the irreversibility of solution field $u$ with respect to the coordinate $\beta_k$ is characterized by
\begin{equation}
    s_k\frac{\partial u}{\partial \beta_k}(\bm{\beta}) \geqslant 0, \quad \forall \bm{\beta} \in \mathcal{D},
    \label{eq:irreversibility}
\end{equation}
where $s_k \in \{+1, -1\}$ is directional symbol indicating \emph{forward} irreversibility ($s_k=+1$) or \emph{backward} irreversibility ($s_k=-1$). 

We introduce an \emph{irreversibility measure} to quantify the degree to which the neural network solution violates the irreversibility constraints, which is zero when the constraints are satisfied and positive when they are violated. For a neural network approximation $\hat{u}\colon \mathcal{D} \to \mathbb{R}$ with parameters $\bm{\theta}$, the pointwise irreversibility violation measure is defined as:
\begin{equation}
    V_k(\boldsymbol{\beta};\bm{\theta})
    = \max\!\left(0,\; - s_k \frac{\partial \hat{u}}{\partial \beta_k}(\boldsymbol{\beta};\bm{\theta})\right).
    \label{eq:irreversibility_measure}
\end{equation}

For computational implementation, the \texttt{max} operation can be realized using the \texttt{ReLU} activation function \cite{agarapDeepLearningUsing2019} (or other smooth approximations such as \texttt{Softplus} \cite{zhengImprovingDeepNeural2015} and \texttt{Swish} \cite{ramachandranSwishSelfGatedActivation2017} if required), which is differentiable almost everywhere (except at zero for \texttt{ReLU}) and fully compatible with gradient-based optimization methods. In this way, Equation~\eqref{eq:irreversibility_measure} can be reformulated as
\begin{equation}
    V_k (\bm{\beta}; \bm{\theta}) = \text{ReLU}\left(s_k\frac{\partial \hat{u}}{\partial \beta_k}(\bm{\beta}; \bm{\theta})\right).
    \label{eq:relu_violation}
\end{equation}

Building upon the defined measure, we construct the irreversibility regularization functional by penalizing violations at selected collocation points within the domain, which are sampled in the same manner as those used for evaluating the PDE residuals in the conventional PINN. Let $\{\bm{\beta}_{\text{irr}}^j\}_{j=1}^{N_{\text{irr}}} \subset \mathcal{D}$ denote the collocation points where the irreversibility constraints are enforced. The regularization functional is then defined as:
\begin{equation}
    \mathcal{L}_{\text{irr}}^k(\bm{\theta}) = \frac{1}{N_{\text{irr}}} \sum_{j=1}^{N_{\text{irr}}} V_k(\bm{\beta}_{\text{irr}}^j; \bm{\theta}),
    \label{eq:irreversibility_loss}
\end{equation}
which provides a differentiable penalty term that can be seamlessly integrated into PINN training procedures.


\subsection*{Benchmark evaluation}

We evaluate the robustness and efficiency of the proposed regularization strategy (denoted as IRR-PINN) through five benchmarks covering traveling-wave propagation, combustion, ice melting, corrosion, and crack growth, and compare the results against those obtained using a conventional PINN and FEM/analytical reference solutions. These physical problems fall broadly into two categories. The first comprises gradient-driven transport processes, characterized by irreversibility along a spatial coordinate $x_m$ ($m \in \{1, 2, \ldots, d\}$) termed \emph{directional} irreversibility. The second involves interfacial evolution phenomena such as corrosion, phase transition, and cracking, referred to as \emph{dissipative} irreversibility corresponding to irreversibility along the temporal coordinate $\beta_k = t$.  

All governing equations for benchmark tests are given in the \texttt{Methods} section. The comparison is carried out at two complementary levels. First, we examine accuracy-related metrics, in particular the $L^2$ error relative to the reference solution. Second, we evaluate the degree to which the predicted fields respect the underlying physical laws, including whether local violations of the irreversibility condition occur and how the irreversibility loss evolves during training. Table \ref{tab:benchmark_results} summarizes the benchmark results with and without irreversibility regularization. Across all benchmarks considered, IRR-PINN consistently achieves substantially lower $L^2$ errors than the conventional PINN, while maintaining the same computational cost (see Section \ref{sec:computational_cost} in the Supplementary Information). This demonstrates that enforcing irreversibility as a soft constraint yields a marked improvement in solution fidelity without incurring additional overhead.

\begin{table}[htbp]
    \centering
    \caption{Summary of benchmark results obtained using IRR-PINN and conventional PINN.}
    \label{tab:benchmark_results}
    \begin{tblr}{
        width=\textwidth,
        colspec={Q[l,1]Q[c,1]Q[c,1]Q[c,1]},
        rowspec={Q[m]Q[m]Q[m]Q[m]Q[m]},
        row{1}={font=\bfseries}}
        \hline
        \SetCell[r=2,c=1]{m} Benchmark test &
        \SetCell[r=2,c=1]{m} Irreversibility type &
        \SetCell[r=1,c=2]{c} Relative $L^2$ error (in $\%$) & \\ \cline{3-4}
        & & IRR-PINN & Conventional PINN \\
        \hline
        Traveling wave propagation & Directional & 0.716 & 100 \\
        Steady combustion          & Directional & 0.464 & 54.9 \\
        Ice melting                & Dissipative & 0.164 & 0.696 \\  
        Corrosion modeling         & Dissipative & 0.118 & 4.07 \\   
        Crack growth       & Dissipative & 2.15  & 7.28 \\
        \hline
    \end{tblr}
\end{table}

\subsection*{Gradient-driven transportation: directional irreversibility}

In gradient-driven transport, fluxes naturally flow from high to low potential, setting a preferred direction for how fronts advance. Once this movement begins, the gradients cannot spontaneously rebuild, giving the process a built-in spatial (directional) irreversibility. We use two representative benchmarks to examine how incorporating this directional irreversibility influences PINN training.

\noindent \textbf{A. Traveling-wave propagation.} We begin by assessing the proposed method using a one-dimensional traveling-wave propagation problem governed by the Fisher-type reaction-diffusion equation. This equation features a balance between nonlinear reaction and diffusive spreading, giving rise to stable traveling fronts. Such models are widely used to describe physical, biological, and chemical systems that combine auto-catalytic growth with diffusive transport \cite{belgacem2012identifiability}, including population invasion, combustion, and tumor progression.

We work on a spatial domain $\Omega \subset [-20,20]$ m and a temporal window $T \subset [0,20]$ s, with the system initiated by a localized Gaussian distribution $u(x, 0) = \exp\left(-x^2\right)$ and held at zero at the domain boundaries. In this setting, the resulting wave fronts exhibit an inherent directional irreversibility: once the front advances, it cannot spontaneously recede. For the given Gaussian initial profile, two fronts emerge from $x_0=0$ and propagate outward in the $\pm x$ directions. Consequently, the solution $u(x, t)$ must satisfy opposite irreversibility constraints on the left and right halves of the domain, which can be unified into a single spatial irreversibility regularization term
\begin{equation}
    \mathcal{L}_{\text{irr}}^x(\bm{\theta}) = \frac{1}{N_{\text{irr}}} \sum_{j=1}^{N_{\text{irr}}} \text{ReLU}\left( \frac{x_{\text{irr}}^j}{\left\lvert x_{\text{irr}}^j\right\lvert + \epsilon_x} \cdot \frac{\partial \hat{u}}{\partial x}(x_{\text{irr}}^j, t_{\text{irr}}^j; \bm{\theta})\right),
    \label{eq:travelling_wave_irreversibility}
\end{equation}%
with $\epsilon_x>0$ being a small constant for numerical stability.

\begin{figure}[ht]
    \centering
    \includegraphics[width=\textwidth]{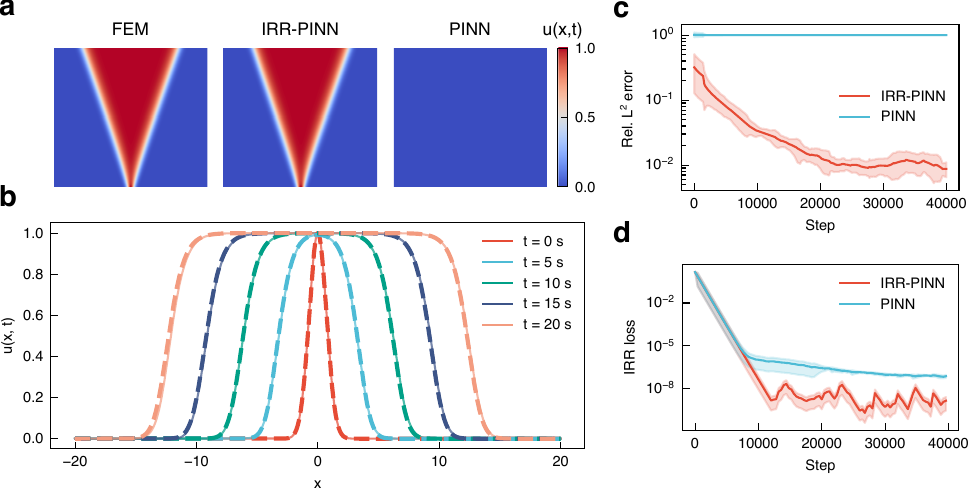}
    \caption{
        \textbf{Traveling wave propagation}. 
        (\textbf{a}) Traveling-wave solution fields computed using the finite element method (FEM), the proposed irreversibility-regularized PINN (IRR-PINN), and a conventional PINN. These abbreviations (FEM, IRR-PINN, PINN) are used throughout the paper.
        (\textbf{b}) Spatial distributions of solution profiles at different time points predicted by IRR-PINN (solid lines) compared with FEM reference solutions (dashed lines), showing excellent agreement. 
        (\textbf{c-d}) Training histories of the relative $L^2$ error and the spatial irreversibility loss for IRR-PINN and conventional PINN. IRR-PINN consistently exhibits superior accuracy and stability across both metrics.
    }
    \label{fig:travellingwave}
\end{figure}

We benchmark the performance of IRR-PINN in this problem by comparing it with predictions from a conventional PINN and with FEM reference solutions. Figure~\ref{fig:travellingwave} summarizes all results. Panel (a) shows the solution fields obtained using the three approaches. IRR-PINN accurately reconstructs the traveling-wave dynamics and closely matches the FEM reference. This agreement is further quantified in panel (b), which compares the solutions at several time points and shows near-perfect overlap between IRR-PINN and FEM. In contrast, the conventional PINN fails to capture the wave propagation and instead predicts an almost flat field near zero across the entire domain. This failure originates from the early stages of training, during which the network is unable to form the sharp front required for gradient-driven propagation. However, when the irreversibility constraint is imposed, this essential physical feature emerges naturally.

Panels (c) and (d) present the training histories of the relative $L^2$ error and the spatial irreversibility loss $\mathcal{L}_{\text{irr}}^x$. Both metrics decrease rapidly to nearly zero for IRR-PINN, remaining several orders of magnitude smaller than those of the conventional PINN. Notably, the conventional PINN maintains a relative $L^2$ error close to $100\%$, indicating its failure to learn the correct solution. Together, these results demonstrate that respecting physical irreversibility is crucial for accurately modeling gradient-driven processes such as traveling-wave propagation.

\noindent \textbf{B. Steady combustion.} We next examine a simple yet representative benchmark that exhibits clear directional irreversibility: steady premixed combustion in one dimension. The goal of this test is to demonstrate that IRR-PINN can markedly improve predictive accuracy even for steady problems with relatively simple ODE. We consider a freely propagating premixed flame, in which a flame front travels unidirectionally through the domain. At the inlet ($x=0$), the inflow temperature $T_\text{in}=298\;\mathrm{K}$ and its gradient
$\left(\mathrm{d}T / \mathrm{d}x\right)_\text{in}=1.0\times 10^5\;\mathrm{K/m}$ are prescribed, and the domain length is $L=1.5\times 10^{-3}\;\mathrm{m}$. Because the flame front advances only in the positive $x$ direction, the temperature field must increase monotonically along the flow, reflecting the intrinsic forward irreversibility of the combustion process. 

\begin{figure}[ht]
    \centering
    \includegraphics[width=\textwidth]{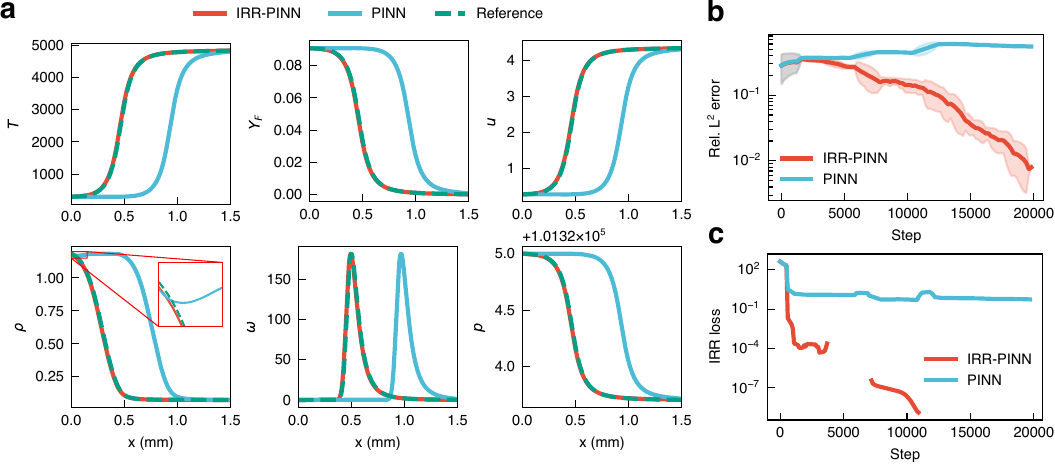}
    \caption{
        \textbf{Steady combustion}.
        (\textbf{a}) Solution fields of temperature $T$ and other derived variables predicted by the IRR-PINN, conventional PINN, and FEM for reference. IRR-PINN demonstrates excellent agreement with FEM, whereas the conventional PINN exhibits substantial deviations.
        (\textbf{b-c}) Training histories of the relative $L^2$ error and the spatial irreversibility loss predicted by both IRR-PINN and conventional PINN. The blank regions in (\textbf{c}) correspond to zero values under the logarithmic scale. IRR-PINN consistently achieves higher accuracy and improved stability across both metrics.
    }
    \label{fig:combustion}
\end{figure}

Figure~\ref{fig:combustion}a compares the temperature field and several derived quantities predicted by IRR-PINN and a conventional PINN against FEM reference solutions. IRR-PINN shows excellent agreement with FEM across all variables, whereas the conventional PINN exhibits large deviations. The difference is especially noteworthy in the gas-density profile $\rho$, which should decrease monotonically along the flame direction but displays non-physical oscillations near the inlet when predicted by the conventional PINN, as highlighted in the magnified inset in Figure~\ref{fig:combustion}a. Such violations of the irreversible structure likely contribute to its overall poor performance. Finally, panels (b) and (c) show the training histories of the relative $L^2$ error of temperature field and the irreversibility loss $\mathcal{L}_{\text{irr}}$. For IRR-PINN, both quantities decrease steadily to values below $0.5\%$, whereas the conventional PINN stagnates at much higher errors. These results highlight the importance of enforcing physical irreversibility, not only to ensure physically meaningful solutions but also to stabilize the training process.

\subsection*{Interfacial evolution: dissipative irreversibility}
Interfacial evolution is a widespread and visually striking phenomenon in nature, governing a broad range of physical processes and involving the interplay of multiple coupled fields. A commonly used approach for modeling this phenomenon is the so-called phase field method \cite{biner2017programming}, in which a smooth order parameter $\phi$ captures the motion and transformation of interfaces. A defining feature of such systems is their strong temporal irreversibility: once an interface forms or advances, it cannot spontaneously retreat without external driving forces. To evaluate the robustness of our irreversibility regularization, we apply IRR-PINN to three representative interfacial problems: ice melting, corrosion, and crack propagation.

\noindent \textbf{A. Ice melting.} We begin our study of interfacial evolution with a simple yet representative example: the melting of a spherical ice inclusion of initial radius $R_0$. Under a uniformly elevated temperature field, the solid-liquid interface retreats smoothly toward the center, and the melting front shrinks linearly in time as $R(t) = R_0 - \lambda t$, where $\lambda$ denotes the melting rate. This process is well described by a single \emph{Allen--Cahn}-type phase field equation, and the melted region ($\phi=-1$) cannot spontaneously revert to solid ($\phi=1$), making the problem an ideal benchmark for evaluating IRR-PINN. In this setting, the irreversibility-regularized loss term can be formulated as
\begin{equation}
    \mathcal{L}_{\text{irr}}^t(\bm{\theta}) = \frac{1}{N_{\text{irr}}} \sum_{j=1}^{N_{\text{irr}}} \text{ReLU}\left(\frac{\partial \hat{\phi}}{\partial t}(\bm{x}_{\text{irr}}^j, t_{\text{irr}}^j; \bm{\theta})\right).
    \label{eq:melt_irreversibility}
\end{equation}

\begin{figure}[ht]
    \centering
    \includegraphics[width=\textwidth]{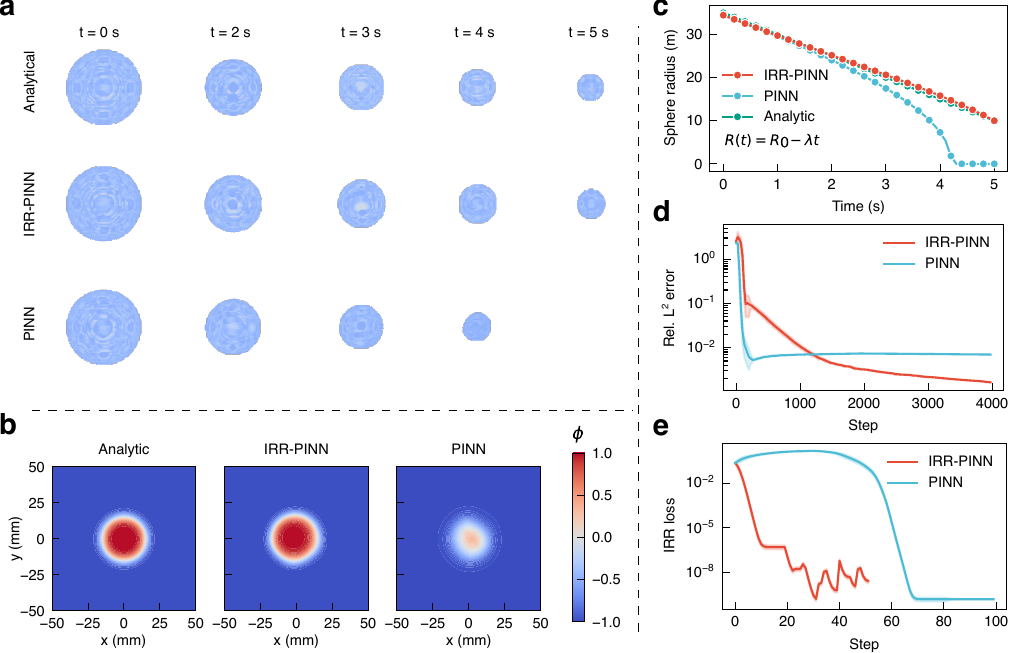}
    \caption{
        \textbf{Ice melting}.
        (\textbf{a}) Phase field solution $\phi$ at different time points predicted by IRR-PINN and conventional PINN, compared with the analytical solution. IRR-PINN closely follows the analytical interface position throughout the simulation, whereas the conventional PINN progressively deviates from the reference. Only the interfacial region ($0.5 \leq \phi \leq 0.5$) is shown for visual clarity.
        (\textbf{b}) Cross-section of $\phi$ at $z=0$ and $t=4$ s, again showing excellent agreement between IRR-PINN and the analytical solution, in contrast to the conventional PINN.
        (\textbf{c}) Temporal evolution of the melting radius predicted by IRR-PINN and the conventional PINN compared with the analytical law. IRR-PINN accurately reproduces the linear retreat of the melting front, while the conventional PINN yields a non-physical, non-linear trajectory. 
        (\textbf{d-e}) Training histories of the relative $L^2$ error of $\phi$ and the irreversibility loss for both IRR-PINN and conventional PINN. Although the conventional PINN shows an initial reduction in error, it stagnates thereafter, whereas IRR-PINN exhibits consistent convergence toward very small errors.
    }
    \label{fig:icemelting}
\end{figure}

We model a three-dimensional melting problem in a cubic domain $\Omega = [-50, 50]^3\;\mathrm{mm}$ over a temporal domain $T = [0, 5]\;\mathrm{s}$. Figure~\ref{fig:icemelting} summarizes the performance of IRR-PINN and conventional PINN against the analytical melting law $R(t) = R_0 - \lambda t$. Panels (a) and (b) show the predicted phase field variable $\phi$, which marks the solid-liquid interface: panel (a) presents three-dimensional snapshots at several time points, and panel (b) shows a cross-section at $z=0$ and $t=4$ s. These contours clearly illustrate that IRR-PINN reproduces the analytical interface position with high fidelity across all times. This agreement is quantified in Figure~\ref{fig:icemelting}c, where IRR-PINN accurately captures the linear retreat of the melting front and its constant rate. By contrast, the conventional PINN rapidly departs from the expected linear trend and predicts a nonphysical melting trajectory. Panels (d-e) further compare the relative $L^2$ error of $\phi$ and the irreversibility loss during training. IRR-PINN converges to a maximum relative error of only $0.22\%$, whereas the conventional PINN begins to drift markedly after the early stages (approximately $t<2\;\mathrm{s}$) due to its large irreversibility loss in the absence of the soft constraint. These results highlight the crucial role of respecting hidden temporal irreversibility in training PINNs for interfacial evolution problems.

\noindent \textbf{B. Corrosion modeling.} Building on the successful simulation of the ice-melting problem, we next evaluate IRR-PINN in a more complex setting: pitting corrosion. Pitting corrosion is a long-standing challenge in many engineering applications exposed to aggressive environments, and accurate prediction of its evolution is critical for ensuring the safety and durability of structural materials and components. The phenomenon is modeled using two primary variables: a phase field $\phi$ that tracks the advancing corrosive interface through an \emph{Allen--Cahn}-type equation, and a normalized concentration $c$ that represents the diffusion of dissolved metal ions and follows a \emph{Cahn--Hilliard}-type equation. The strong coupling between these two equations makes the problem particularly difficult for the conventional PINN \cite{chenPFPINNsPhysicsinformedNeural2025}, thereby providing a stringent test for the proposed irreversibility-regularized strategy. 

\begin{figure}[H]
    \centering
    \includegraphics[width=\textwidth]{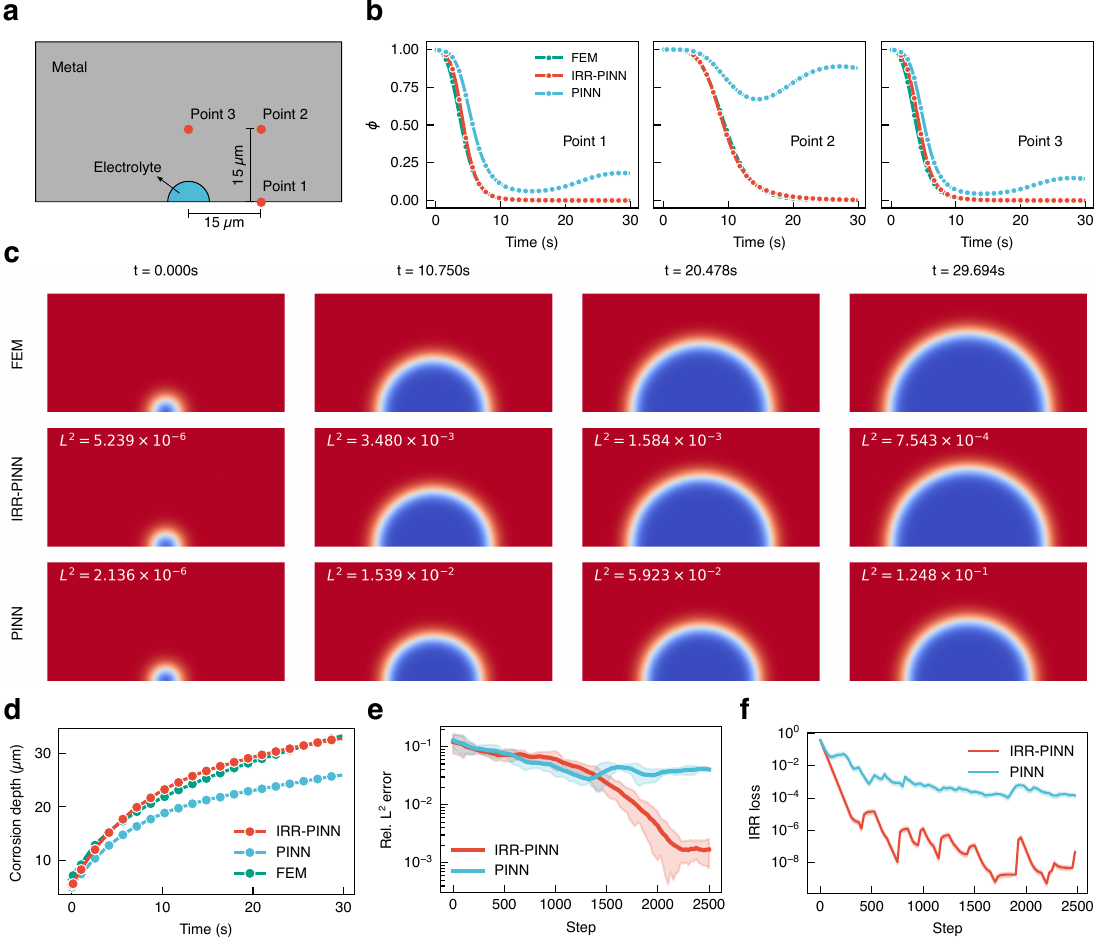}
    \caption{
        \textbf{Corrosion modeling}.
        (\textbf{a}) Schematic of the two-dimensional semi-circular pit-growth problem with an initial pit introduced at the center of the bottom boundary. Red markers indicate representative monitoring locations used to track the temporal evolution of corrosion.
        (\textbf{b}) Temporal evolution of $\phi$ at the three monitoring points, showing that the conventional PINN violates dissipative irreversibility, in contrast to IRR-PINN and FEM.
        (\textbf{c}) Solution fields of phase field $\phi$ at different time points predicted by IRR-PINN and conventional PINN, compared with FEM reference solutions. Red denotes the metal phase $\phi=1$, and blue denotes the corroded region $\phi=0$. The conventional PINN exhibits noticeable deviations across all times.
        (\textbf{d}) Time-dependent evolution of the maximum corrosion depth predicted by the three approaches. IRR-PINN remains in excellent agreement with the FEM reference, whereas the conventional PINN substantially underestimates pit growth due to non-physical reversals of the interface.
        (\textbf{e-f}) Training histories of the relative $L^2$ error and the irreversibility loss for highlighting the performance of the IRR-PINN compared to the conventional PINN framework.
    }
    \label{fig:corrosion}
\end{figure}

We consider a two-dimensional semi-circular pit growth problem, as shown in Figure~\ref{fig:corrosion}a. The spatial domain is $\Omega = [-50, 50]\;\mathrm{\mu m} \times [0, 50]\;\mathrm{\mu m}$ and the temporal window is $T = [0, 30]\;\mathrm{s}$. A small initial pit is introduced at the center of the bottom boundary with $\phi=c=0$ to initiate corrosion. As in the ice melting example, pitting corrosion is inherently irreversible: the transformation progresses only from metal ($\phi=1$) toward electrolyte ($\phi=0$) and cannot spontaneously reverse. Accordingly, the phase field variable $\phi$ must satisfy an irreversibility constraint along the temporal dimension. To directly assess potential violations, three monitoring points, marked in red in Figure~\ref{fig:corrosion}a, are selected for comparison with and without the irreversibility constraint. Not surprisingly, as shown in Figure~\ref{fig:corrosion}b, the conventional PINN without irreversibility constraint increases in the phase field variable $\phi$ at several time points, indicating a spontaneous and non-physical reformation of the metal phase. In contrast, the IRR-PINN solution follows the same monotonic decrease of $\phi$ as the FEM reference, correctly preventing the interface from recovering once corrosion has occurred.

Figure~\ref{fig:corrosion}c compares the phase field variable $\phi$, which tracks the corrosion front, predicted by IRR-PINN and a conventional PINN against the FEM reference at several time points. A quantitative comparison of the maximum pit depth is provided in Figure~\ref{fig:corrosion}d. Across all examined times, IRR-PINN closely matches the FEM solution, accurately capturing both the pit morphology and its temporal evolution. In conjunction with Figure~\ref{fig:corrosion}b, it is clear that this superior performance arises from the beneficial effect of the imposed irreversibility constraint. This accuracy is further reflected in a maximum relative $L^2$ error below $0.35\%$, as shown in Figure~\ref{fig:corrosion}e, and a substantially smaller irreversibility loss in Figure~\ref{fig:corrosion}f. In contrast, the conventional PINN shows pronounced deviations from the FEM reference, consistently underestimating the pit depth as corrosion progresses. 

\noindent \textbf{C. Crack growth.} We now close the benchmark tests by simulating a crack propagation example modeled within the phase field fracture paradigm. As with other interfacial phenomena, fracture exhibits intrinsic irreversibility: once a crack initiates, the damaged region ($\phi=1$) cannot heal back to the intact state ($\phi=0$). Unlike melting or corrosion, however, phase field fracture cannot be driven solely by prescribed initial or boundary conditions; crack growth is driven by the strain energy released under external loading. As the load increases, the system undergoes a sharp, highly nonlinear transition at the crack-nucleation threshold, which is a feature that the conventional PINN struggle to reproduce. For this reason, energy-based approaches such as the Deep Ritz Method \cite{manavPhasefieldModelingFracture2024b} are commonly used. Here, we demonstrate that our irreversibility-regularized strategy enables PINNs to handle this challenging problem effectively.

A key requirement in phase field fracture is the Karush--Kuhn--Tucker (KKT) condition, which enforces damage irreversibility and turns the evolution problem into an inequality-constrained variational formulation. In traditional finite element time stepping, this constraint is enforced either through local constrained minimization at each step \cite{Wu2017,Feng2021}  or by introducing a history-based driving force \cite{Miehe2010a,Kristensen2021}. Both strategies rely on incremental updates and are therefore incompatible with a global space-time PINN formulation. Instead, we impose the KKT condition directly through a pointwise residual combined with the dissipative irreversibility regularization. Implementation details are provided in Section \ref{sec:PFF} in the Supplementary Information.

\begin{figure}[ht]
    \centering
    \includegraphics[width=\textwidth]{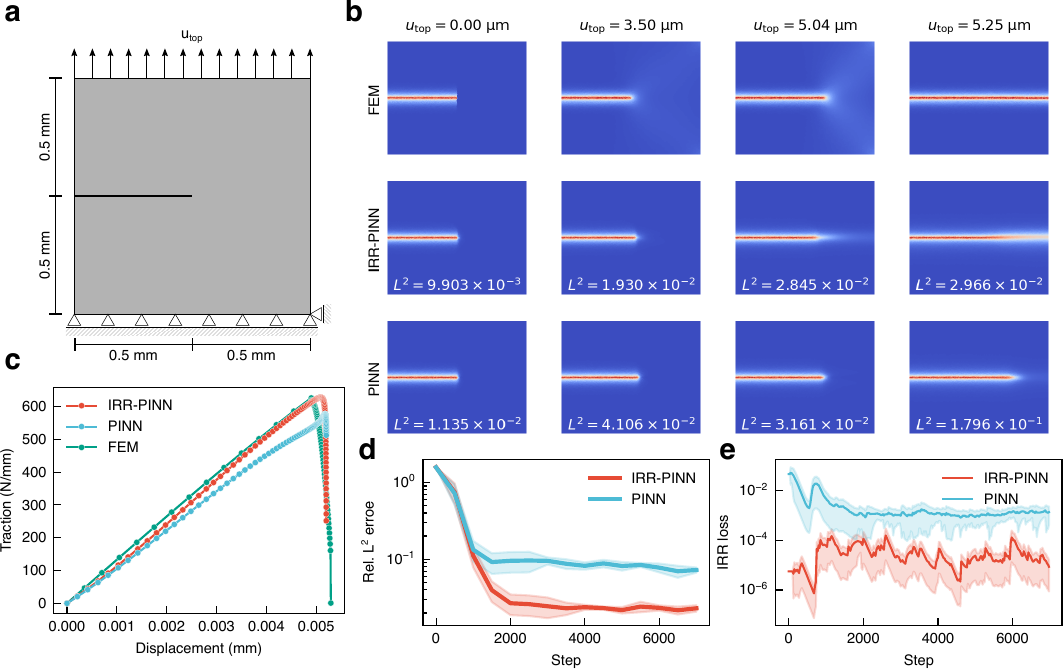}
    \caption{
        \textbf{Crack growth}.
        (\textbf{a}) Geometric setup of a two-dimensional single-edge notched tension test. An initial crack of length $0.5\;\mathrm{mm}$ is prescribed along the mid-height of the left edge. A time-dependent vertical displacement $u_\text{top}(t)$ is applied to the top boundary to drive crack propagation.
        (\textbf{b}) Phase field solution $\phi$ at different loading stages predicted by IRR-PINN and a conventional PINN, compared with FEM reference solutions. Only IRR-PINN successfully captures the correct crack-propagation rate.
        (\textbf{c}) Load-displacement curves predicted by IRR-PINN, conventional PINN, and FEM. The conventional PINN deviates from the reference response gradually, resulting in inaccurate predictions of both crack nucleation and subsequent propagation.
        (\textbf{d-e}) Training histories of the relative $L^2$ error and  the irreversibility loss for IRR-PINN and the conventional PINN. IRR-PINN achieves substantially better accuracy and physical consistency across both metrics.
    }
    \label{fig:fracture}
\end{figure}

To validate the proposed framework, we examine a classical fracture benchmark: a two-dimensional single-edge notched tension specimen, illustrated in Figure~\ref{fig:fracture}a. Crack nucleation and propagation are driven by a time-dependent vertical displacement applied to the top boundary. To emphasize the nonlinear fracture response rather than the initial elastic stage, we prescribe a smooth loading protocol that rapidly ramps up to a target displacement and then remains constant
\begin{equation}
    u_\text{top}(t) = u_r \cdot \frac{\tanh\left(\alpha t\right)}{\tanh(\alpha)}, \quad t \in [0, 1].
    \label{eq:fracture-displacement-control}
\end{equation}
A pre-existing crack is introduced through an initial phase field profile that sharply localizes the damaged region along half of the left edge. This configuration produces a clear crack nucleation event followed by rapid crack growth, providing a stringent test of the ability of IRR-PINN to capture strongly nonlinear and irreversible fracture mechanics.

Figure~\ref{fig:fracture}b compares the phase field variable $\phi$ predicted by IRR-PINN and the conventional PINN against the FEM reference solutions at several loading stages. When the applied load is small and the response remains nearly linear, both models produce reasonably accurate results. However, as the crack begins to grow in a brittle manner and the nonlinear response becomes dominant, the conventional PINN significantly underestimates the crack propagation rate. In contrast, IRR-PINN accurately predicts the crack length across all loading levels and closely matches the FEM reference. This improvement is further reflected in the force-displacement curves shown in Figure~\ref{fig:fracture}c. The conventional PINN captures only the very early elastic regime before deviating sharply from the FEM response, whereas IRR-PINN faithfully reproduces the entire curve, including the peak load associated with crack nucleation and the subsequent post-peak softening corresponding to crack propagation. To the best of our knowledge, this represents the first successful simulation of phase field fracture within the PINN framework.

Finally, panels (d-e) show the evolution of the relative $L^2$ error and the irreversibility loss $\mathcal{L}_{\text{irr}}^t$ during training. IRR-PINN consistently outperforms the conventional PINN in both metrics, achieving near-zero irreversibility violations while maintaining a relative error around $2.0\%$. These quantitative results highlight the crucial role of enforcing irreversibility in enabling accurate and physically consistent fracture simulations.

\section*{Discussion}
Physics-informed neural networks (PINNs) have gained significant attention and have been deployed across a broad spectrum of physical problems. Yet questions remain about their ability to faithfully capture all relevant physical laws when used as forward PDE solvers. In this work, we identify a fundamental gap: the conventional PINN often fail to respect \emph{the Second Law of Thermodynamics}, leading to violations of irreversible behavior that is not explicitly encoded in the loss formulation. To address this issue, we introduce a simple, robust, and broadly applicable regularization strategy that enforces irreversible physics as a soft constraint. Two representative forms of irreversibility, namely directional (spatial) and dissipative (temporal) irreversibility, are incorporated through a single additional loss term that augments the conventional PINN formulation.

A notable finding is that these improvements do not increase computational cost. Across all benchmark tests, IRR-PINN required essentially the same training time as the conventional PINN despite the presence of an additional loss term, as shown in Section \ref{sec:computational_cost} in the Supplementary Information. While PINNs are known to be less computationally efficient than classical solvers on small-scale problems, they can potentially provide substantial gains in settings with large domains, extended time horizons, or high-dimensional solution spaces. Our large-scale benchmarks, including ice melting and crack growth, clearly illustrate this trend, with PINN-based models achieving accurate solutions more efficiently than conventional numerical methods.

We further assess and discuss the generalization capacity of IRR-PINN. For each benchmark problem, IRR-PINN and the conventional PINN were trained using the same network architecture and the same set of hyperparameters. Although performance of the conventional PINN can be improved through extensive hyperparameter tuning or through specialized training strategies, our results show that with identical setups, IRR-PINN consistently performs better. Moreover, sensitivity analyses provided Figure \ref{fig:travellingwave-hyperparameter-sensitivity} in Supplementary Information show that IRR-PINN is significantly less sensitive to hyperparameter choices than the conventional PINN. This indicates enhanced robustness and generalization ability as a forward PDE solver. Also, it is worth noting that although the present study focuses on systems governed predominantly by a single form of irreversibility, the proposed regularization can naturally generalize to more complex systems involving multiple sources of irreversible behavior by adaptively placing irreversibility collocation points in regions where violations most strongly affect physical consistency.

Looking ahead, the extension of this approach to physics-informed neural operators (PINOs) provides a promising avenue for modeling irreversible dynamics efficiently across geometries and parameter spaces. Furthermore, integration with generative modeling frameworks such as diffusion models or rectified flow models could improve uncertainty quantification and enable data efficient learning of irreversible processes. These directions point toward scalable and physically-consistent machine learning models for complex real world systems.
In conclusion, IRR-PINN provides a simple and effective paradigm for embedding hidden physical laws within neural network based solvers, and represents an important step toward more reliable and physically-grounded scientific machine learning.

\section*{Methods}

The preceding results demonstrate that IRR-PINN can efficiently handle a wide range of physical problems. We now provide an overview of the computational framework and the governing equations underlying each benchmark. Additional details, including implementation specifics, hyperparameter settings, mathematical derivations, and numerical procedures, are provided in the Supplementary Information.

\subsection*{Baseline physics-informed neural networks}

We construct a strong baseline PINN framework by extending the original formulation \cite{raissiPhysicsinformedNeuralNetworks2019} with a suite of state-of-the-art techniques. These enhancements are integrated to maximize training stability and solution accuracy, thereby ensuring a high-quality baseline for evaluating the proposed irreversibility-regularization strategy. Key components of the baseline PINN framework are summarized below.

\noindent \textbf{Neural network architecture.} We employ fully connected networks as the backbone architecture, selecting from standard MLPs, ResNets \cite{he2016deep}, or modified MLPs \cite{wangUnderstandingMitigatingGradient2021} based on problem complexity. The modified MLP, which uses gating mechanisms to mitigate gradient pathologies, is defined as:
\begin{subequations}
    \begin{gather}
    \bm{U} = \alpha\left(
        \bm{W}_u \bm{z}^{(0)} + \bm{b}_u
    \right), \\
    \bm{V} = \alpha\left(
        \bm{W}_v \bm{z}^{(0)} + \bm{b}_v
    \right), \\
    \hat{\bm{z}}^{(l)} = \alpha\left(
        \bm{W}^{(l)}\bm{z}^{(l-1)} + \bm{b}^{(l)}
    \right), \\
    \bm{z}^{(l)} = \hat{\bm{z}}^{(l)} \odot \bm{U} + \left(1 - \hat{\bm{z}}^{(l)}\right) \odot \bm{V}, \quad l = 1, 2, \ldots, L,
    \end{gather}
\end{subequations}
where $\bm{z}^{(0)}$ denotes the input, $\bm{z}^{(l)}$ the output of the $l$-th hidden layer, $\alpha(\cdot)$ the activation function, and $\odot$ element-wise multiplication.

\noindent \textbf{Random Fourier feature embedding.} To mitigate spectral bias and improve the representation of high-frequency solution components, we apply Random Fourier Feature Embedding (FFE) \cite{wangEigenvectorBiasFourier2021} to the input coordinates. For an input vector $\bm{v}$, the FFE mapping is given by:
\begin{equation}
    \mathcal{F}(\bm{v}) = \left[\begin{matrix}
        \cos\left(\bm B\bm v\right)\\
        \sin\left(\bm B\bm v\right)
    \end{matrix}
    \right],
\end{equation}
where $\bm{B} \in \mathbb{R}^{m \times n}$ is a random projection matrix with entries sampled from a Gaussian distribution $\mathcal{N}(0, \sigma^2)$, and $\sigma$ is a hyperparameter governing the frequency scale.

\noindent \textbf{Causal training.} For time-dependent problems, we adopt a causal training scheme \cite{wangRespectingCausalityTraining2024} to enforce temporal causality. The temporal domain is partitioned into $N_t$ segments, and the PDE residual loss for each segment is weighted according to the cumulative loss from preceding segments. The causal weight for the $i$-th segment is defined as:
\begin{equation}
    w_\text{causal}^i = \exp\left(
        -\epsilon_c \sum_{j=0}^{i-1} \mathcal{L}_g^j(\bm{\theta})
    \right), \quad i = 0, 1, \ldots, N_t,
\end{equation}
where $\epsilon_c$ controls the strength of the causal weighting, and $\mathcal{L}_g^j$ denotes the PDE residual loss for the $j$-th segment.

\noindent \textbf{Staggered training scheme.} In scenarios involving coupled PDEs (e.g., phase field fracture) or multi-objective optimization (e.g., combustion with eigenvalue estimation), gradient conflicts \cite{wangGradientAlignmentPhysicsinformed2025a} can hinder convergence and degrade accuracy. To address this, we employ a staggered training strategy \cite{chenSharpPINNsStaggeredHardconstrained2025} that updates network parameters associated with each equation or objective in an alternating fashion. For eigenvalue problems, we further stabilize training by fixing the eigenvalue while updating network parameters, and vice versa.

\noindent \textbf{Gradient-normalized loss weighting.} To balance multiple loss terms adaptively, we implement a gradient-normalized loss weighting scheme \cite{wangExpertsGuideTraining2023}. The weight for each loss term is computed based on the relative magnitudes of their gradients. For a set of loss terms $\{\mathcal{L}_j\}_{j \in \mathcal{J}}$, the weights at the $s$-th training iteration are updated as:
\begin{subequations}
    \begin{gather}
        \hat w_j^{(s)} = \frac{\displaystyle\sum_{j\in\mathcal{J}} \| \nabla_{\bm\theta}\mathcal{L}_j^{(s)} \|}{\| \nabla_{\bm\theta}\mathcal{L}_j^{(s)} \|},
        \quad s \geqslant 1, \quad \forall j\in\mathcal{J}, \\
        w_j^{(s)} = \alpha_w \cdot \hat w_j^{(s-1)} + (1-\alpha_w) \cdot w_j^{(s)}, \\
        w_j^{(0)} = 1,
    \end{gather}%
\end{subequations}
where $\alpha_w \in [0,1)$ is a smoothing parameter. It should be noted that this adaptive weighting strategy is applied to all loss terms, including the irreversibility regularization introduced in this work.

\subsection*{Governing equations for the benchmark tests}

\noindent\textbf{Traveling-wave propagation.} The one-dimensional traveling-wave propagation is defined as
\begin{equation}
    \frac{\partial u}{\partial t}  = \frac{\partial^2 u}{\partial x^2} +  u(1-u).
\end{equation}

\noindent\textbf{Steady combustion.} The steady combustion process is governed by an ordinary differential equation (ODE), and a simplified form of the governing equation for this problem is given by \cite{wuFlamePINN1DPhysicsinformedNeural2025}
\begin{equation}
    \rho_\text{in}s_\text{L}c_p\frac{\mathrm{d}T}{\mathrm{d}x} - \lambda \frac{\mathrm{d}^2 T}{\mathrm{d}x^2} = -\omega q_F, \label{eq:combustion-pde}
\end{equation}
where $T$ is the temperature field to be solved for and $s_\text{L}$ is the laminar flame speed, which is treated as an unknown eigenvalue in this problem. Additional relations required to close the model are provided in the Supplementary Information. 

It is important to note that the eigenvalue $s_\text{L}$ is not known in advance and must be identified as part of the solution. To address this, we treat $s_\text{L}$ as an additional trainable parameter within the neural network and optimize it jointly with other network parameters $\bm{\theta}$. 

\noindent\textbf{Ice melting.} The ice melting process is decribed by a single \emph{Allen--Cahn} equation admitting an analytical solution. An idealized, constant temperature field $T\equiv 1$ is prescribed to drive the phase transition, such that the governing equation can be expressed as 
\begin{equation}
    \frac{\partial \phi}{\partial t} = M\left(
        \Delta \phi - \frac{F'(\phi)}{\ell^2}  
    \right)- \lambda\frac{\sqrt{2F(\phi)}}{\ell},
    \label{eq:ice-melting-pde}
\end{equation}
where $F(\phi) = \dfrac{1}{4}(\phi^2-1)^2$ is the double-well potential, $M$ is a constant mobility, $\ell$ is the interface thickness, and $\lambda$ is a melting rate parameter. The initial phase field distribution is given by
\begin{equation}
    \phi(x, y, z, 0) = \tanh\left(\frac{R_0 - \sqrt{x^2 + y^2 + z^2}}{\sqrt{2}\ell}\right). \label{eq:ice-melting-initial}
\end{equation}

\noindent\textbf{Corrosion modeling.} The evolution of pitting corrosion is simulated using a KKS (Kim-Kim-Suzuki)-based phase field model \cite{maiPhaseFieldModel2016,cuiPhaseFieldFormulation2021}, in which the metal-electrolyte interface is explicitly represented by a phase field variable $\phi$ that transitions smoothly from 1 (metal) to 0 (electrolyte) following the \emph{Allen--Cahn} equation. In addition, a diffusion-type \emph{Cahn--Hilliard} equation is also employed to describe the transport of the normalized metal ion $c$ and to distinguish between activation-controlled and diffusion-controlled corrosion processes \cite{maiPhaseFieldModel2016}. The governing equations of this problem are presented directly below, and a more detailed formulation can be found in Refs. \cite{chenPFPINNsPhysicsinformedNeural2025,chenSharpPINNsStaggeredHardconstrained2025}.
\begin{subequations}
    \begin{align}
        \text{Cahn--Hilliard: } & \frac{\partial c}{\partial t}- 2\mathcal{A}M \Delta c + 2\mathcal{A}M \left(c_{\mathrm{Se}}-c_{\mathrm{Le}}\right) \Delta h\left(\phi\right) =0 , \label{eq:chcorro} \\
        \text{Allen--Cahn: }    & \frac{\partial \phi}{\partial t}
        -2\mathcal{A}L\left[c-h(\phi)\left(c_{\mathrm{Se}}-c_{\mathrm{Le}}\right)-c_{\mathrm{Le}}\right]\left(c_{\mathrm{Se}}-c_{\mathrm{Le}}\right) h^{\prime}(\phi) +L w_\phi g^{\prime}(\phi) - L \alpha_\phi \Delta \phi =0.
        \label{eq:accorro}
    \end{align}
    \label{eq:corrosion-pde-simplified}
\end{subequations}%
The variables and parameters involved in Equation \eqref{eq:corrosion-pde-simplified} are categorized as follows:
\begin{itemize}
    \item Unknown fields: phase field variable $\phi(\bm x, t)$ and normalized concentration $c(\bm x, t)$;
    \item Derived variables: solid and liquid phase concentrations $c_\text{S}(\bm x, t)$ and $c_\text{L}(\bm x, t)$, with $c_\text{S}(\bm x, t)+c_\text{L}(\bm x, t)\equiv1$;
    \item Material constants: $c_\text{Se}$, $c_\text{Le}$, $\mathcal{A}$, $w$, $\alpha_\phi$, $M$, $L$, which are given in the Supplementary Information.
\end{itemize}

\noindent\textbf{Crack growth.} Phase field fracture involves two strongly coupled fields: the mechanical displacement field $\bm{u}(\bm{x}, t)$, and the phase field variable $\phi(\bm{x}, t)$. The mechanical response of the solid material is described by a degraded linear elastic model, in which the constitutive relationship is modulated by the damage $\phi$. The mechanical equilibrium equation for the displacement field is given by:
\begin{equation}
    \bm\nabla \cdot \left[ g(\phi) \bm \sigma\right] = \bm{0}, \label{eq:fracture-mechanical-equilibrium}
\end{equation}
and the governing equation for crack evolution reads
\begin{equation}
    r=\frac{G_{c}}{\ell} \left(\phi - \ell^2 \nabla^2 \phi\right)+ g^{\prime}(\phi)\,\psi_{0}(\bm{\varepsilon})\geqslant0,
    \label{eq:fracture-driving-force}
\end{equation}
which, together with the irreversibility of the phase field, must satisfy the Karush–Kuhn–Tucker (KKT) conditions:
\begin{equation}
    \dot{\phi}\geqslant0, \quad r\dot{\phi}=0, \quad \text{and}\quad\phi\in[0,1],
    \label{eq:KKT}
\end{equation}
where $g(\phi)=\left(1-\phi\right)^2$ is the degradation function, $\bm \sigma $ is the Cauchy stress tensor, $G_c$ is the critical energy release rate, $\ell$ is the characteristic length scale, and $\psi_{0}(\bm{\varepsilon})$ is the elastic strain energy density.

\putbib[bibfile-cnx,bibfile-ccj,bibfile-irr,bibfile]
\end{bibunit}

\section*{Acknowledgments}

This work was supported in part by National Natural Science Foundation of China (grant number 52478199) and National Key R\&D Program of China (grant number 2021YFF0501003). C.C. additionally acknowledges support from UKRI Horizon Europe Guarantee MSCA Postdoctoral Fellowship (grant EP/Y028236/1). 

\section*{Author contributions statement}

N.C., S.W., and C.C. conceived the project and jointly developed the methodology. N.C. conducted all benchmark computations and completed all coding work. N.C., S.W., and C.C. jointly analyzed the data and interpreted the results. R.M., A.C., and C.C. supervised the project. N.C. and R.M. prepared the original draft. C.C. led the revision of the manuscript with contributions from all authors.

\section*{Code and Reproducibility}

Data and code used in this paper are made freely available at \href{https://github.com/NanxiiChen/irr-pinns}{https://github.com/NanxiiChen/irr-pinns}.  Detailed annotations of the code are also provided.

\section*{Competing interests}

The authors declare no competing interests.

\section*{Additional information}

\textbf{Supplementary information} is available in supplementary file.

\label{MainTextEnd} 
\clearpage
\setcounter{page}{1}
\setcounter{section}{0}
\setcounter{figure}{0}
\setcounter{table}{0}
\setcounter{equation}{0}

\renewcommand{\thepage}{S\arabic{page}}
\renewcommand{\thesection}{S\arabic{section}}
\renewcommand{\thetable}{S\arabic{table}}
\renewcommand{\thefigure}{S\arabic{figure}}
\renewcommand{\theequation}{S\arabic{equation}}

\pagestyle{SI}
\thispagestyle{empty}
\begin{bibunit}[naturemag-doi]
{\LARGE\sffamily\bfseries\noindent
Supplementary Information for\\[2mm]
\textit{Enforcing hidden physics in physics-informed neural networks}
}

\section{Basic formulation of physics-informed neural networks (PINNs)}
\label{sec:basic-pinns-formulation}

Let us first revisit the basic formulation of PINNs \cite{raissiPhysicsinformedNeuralNetworks2019}. PINNs are essentially neural networks that are trained to satisfy both data and physical laws---generally expressed as partial differential equations. By incorporating the residuals of these equations into the loss function, PINNs can effectively learn solutions that adhere to the underlying physics.

Consider a general time-dependent PDE system with an unknown solution $u(\bm x, t, \bm \lambda)$, where $\bm x$ is the spatial coordinate, $t$ is time, and $\bm \lambda$ represents the parameters of the PDE. The PDE system can be expressed as:
\begin{subequations}
    \begin{align}
        \mathcal{G}(\bm{x}, t, u(\bm x, t, \bm \lambda)) &= 0, \quad \bm{x} \in \Omega, t \in [0, T],\\
        \mathcal{B}(\bm{x}, t, u(\bm x, t, \bm \lambda)) &= 0, \quad \bm{x} \in \partial \Omega, t \in [0, T],\\
        \mathcal{I}(\bm{x}, u(\bm x, 0, \bm \lambda)) &= 0, \quad \bm{x} \in \Omega,
    \end{align}
    \label{eq:pde}
\end{subequations}%
where $\mathcal{G}$ is the governing PDE operator, $\mathcal{B}$ and $\mathcal{I}$ represent the boundary and initial conditions, respectively. The solution $u(\bm x, t, \bm \lambda)$ is approximated by a neural network $\mathcal{N}$ with trainable parameters $\bm \theta$ as:
\begin{equation}
    \hat{u}(\bm x, t, \bm \lambda) = \mathcal{N}(\bm x, t; \bm \theta).
\end{equation}
The neural network is trained by minimizing a composite loss function $\mathcal{L}$ that includes contributions from the PDE residuals $\mathcal{L}_g$, boundary conditions $\mathcal{L}_b$, and initial conditions $\mathcal{L}_i$
\begin{equation}
    \mathcal{L} = w_g \mathcal{L}_g + w_b \mathcal{L}_b + w_i \mathcal{L}_i \label{eq:standard-loss-weighted-sum},
\end{equation}
with
\begin{subequations}
    \begin{align}
        \mathcal{L}_g &= \frac{1}{N_g} \sum_{j=1}^{N_g} \left| \mathcal{G}\left(\bm x_g^{j}, t_g^{j}, \hat u\right) \right|^2, \label{eq:loss-governing} \\
        \mathcal{L}_b &= \frac{1}{N_b} \sum_{j=1}^{N_b} \left| \mathcal{B}\left(\bm x_b^{j}, t_b^{j}, \hat u\right) \right|^2, \label{eq:loss-boundary} \\
        \mathcal{L}_i &= \frac{1}{N_i} \sum_{j=1}^{N_i} \left| \mathcal{I}\left(\bm x_i^j, \hat u\right) \right|^2. \label{eq:loss-initial}
    \end{align}
    \label{eq:standard-loss}
\end{subequations}
In this formulation, the sets $\{ \bm x_g^j, t_g^j \}_{j=1}^{N_g}$, $\{ \bm x_b^j, t_b^j \}_{j=1}^{N_b}$ and $\{ \bm x_i^j \}_{j=1}^{N_i}$ represent the space-time coordinates of sampling points corresponding to the differential equations, boundary constraints, and initial conditions, respectively. Effective distribution of these collocation points across the computational domain and its boundaries is essential for training success \cite{wuComprehensiveStudyNonadaptive2023}. The weighting parameters $w_g$, $w_b$, and $w_i$ serve to balance the relative importance of different loss terms and may be prescribed based on prior knowledge or adaptively adjusted throughout the optimization process \cite{wangUnderstandingMitigatingGradient2021a,wangWhenWhyPINNs2022}. Also note that the above PINN formulations for time-dependent problems and their associated loss function definitions can also be compatible with steady-state problems by simply omitting the time variable $t$ and the initial condition term $\mathcal{L}_i$.

\section{Baseline PINN implementation}

This section provides a detailed description of the baseline PINN implementation employed in this study. It should be noted that this implementation serves as a foundational framework upon which our proposed irreversibility regularization method is built. Except for the addition of the irreversibility regularization term in the loss function, all other components and strategies described herein are consistently applied across all benchmark problems to ensure a fair comparison.

\subsection{Neural network architecture}
In this study, we generally utilize the classic ``linear units followed by non-linear activations'' architecture for constructing the backbone neural networks in PINNs. According to the different connecting and stacking patterns of these basic building blocks, we consider three types of neural network architectures: MLP, ResNet, and ModifiedMLP. 

The MLP architecture is the most straightforward one, where each layer is fully connected to the next layer in a sequential manner. The mathematical representation of a standard MLP with $L$ layers can be expressed as:
\begin{equation}
    \bm{z}^{(l)} = \alpha\left(
        \bm{W}^{(l)}\bm{z}^{(l-1)} + \bm{b}^{(l)}
    \right), \quad l = 1, 2, \ldots, L,
\end{equation}
where $\bm{z}^{(l-1)}$ and $\bm{z}^{(l)}$ denote the input and output of the $l$-th layer, respectively; $\bm{W}^{(l)}$ and $\bm{b}^{(l)}$ are the weight matrix and bias vector of the $l$-th layer; and $\alpha(\cdot)$ is a point-wise non-linear activation function. The total trainable parameters of the MLP architecture include all weights and biases across all layers, i.e., $\bm \theta = \{ \bm W^{(l)}, \bm b^{(l)} \}_{l=1}^{L}$.
The ResNet architecture introduces skip connections that allow the input of a layer to bypass one or more layers and be added to the output of a subsequent layer, which helps mitigate the vanishing gradient problem and enables training of deeper networks \cite{he2016deep}, as expressed mathematically:
\begin{equation}
    \bm{z}^{(l)} = \alpha\left(
        \bm{W}^{(l)}\bm{z}^{(l-1)} + \bm{b}^{(l)}
    \right) + \bm{z}^{(l-1)}, \quad l = 1, 2, \ldots, L.
\end{equation}
The trainable parameters remain the same as in the MLP architecture.

The ModifiedMLP architecture incorporates gating mechanisms and residual connections to further enhance the network's ability to capture complex patterns \cite{wangUnderstandingMitigatingGradient2021}. Firstly, two fully connected layers serve as the gating units to modulate the information flow between the hidden layers, as given by:
\begin{subequations}
    \begin{gather}
    \bm{U} = \alpha\left(
        \bm{W}_u \bm{z}^{(0)} + \bm{b}_u
    \right), \\
    \bm{V} = \alpha\left(
        \bm{W}_v \bm{z}^{(0)} + \bm{b}_v
    \right).
    \end{gather}
\end{subequations}
The hidden state $\bm{z}^{(l)}$ is then updated using the weighted sum of the previous hidden state $\bm{z}^{(l-1)}$ and the gating information $\bm{U}$ and $\bm{V}$:
\begin{subequations}
    \begin{gather}
        \hat{\bm{z}}^{(l)} = \alpha\left(
            \bm{W}^{(l)}\bm{z}^{(l-1)} + \bm{b}^{(l)}
        \right), \\
        \bm{z}^{(l)} = \hat{\bm{z}}^{(l)} \odot \bm{U} + \left(1 - \hat{\bm{z}}^{(l)}\right) \odot \bm{V}, \quad l = 1, 2, \ldots, L,
    \end{gather}
\end{subequations}
where $\odot$ denotes element-wise multiplication. The total trainable parameters for the ModifiedMLP architecture are $\bm \theta = \{\bm W_u, \bm b_u, \bm W_v, \bm b_v\}\cup \{  \bm W^{(l)}, \bm b^{(l)} \}_{l=1}^{L}$.

\subsection{Random Fourier feature embedding for spectral bias mitigation}

Spectral bias is a common issue that hiders the ability of neural networks to accurately approximate high-frequency components of the target function \cite{wangWhenWhyPINNs2022,wangEigenvectorBiasFourier2021}, such as the thick interface regions in phase field modeling. To mitigate this issue, we employ the random Fourier feature embedding (FFE) as a preprocessing step prior to feeding the input coordinates into the neural network.

The FFE transforms the original input coordinates $(\bm{x}, t)$ into a higher-dimensional space by applying a simple random Fourier mapping:
\begin{equation}
    \mathcal{F}(\bm{v}) = \left[\begin{matrix}
        \cos\left(\bm B\bm v\right)\\
        \sin\left(\bm B\bm v\right)
    \end{matrix}
    \right],
\end{equation}
where $\bm B \in \mathbb{R}^{m_f\times d}$ is a randomly sampled projection matrix with each entry drawn from a Gaussian distribution $\mathcal{N}(0, \sigma^2)$, $m_f$ is the number of Fourier features, and $d$ is the dimension of the input coordinates, and $\sigma$ is a hyperparameter that controls the frequency range of the Fourier features. The transformed coordinates $\mathcal{F}(\bm{v})$ are then used as the input to the neural network instead of the original coordinates $(\bm{x}, t)$.

For steady-state problems, the input vector is simply $\bm{v} = \bm{x}$, while for time-dependent problems, we apply the FFE separately to the spatial and temporal coordinates and concatenate the results:
\begin{equation}
    \mathcal{F}(\bm{x}, t) = \mathcal{F}_x(\bm{x}) \oplus \mathcal{F}_t(t) = \left[\begin{matrix}
        \cos\left(\bm B_x\bm x\right)\\
        \sin\left(\bm B_x\bm x\right)\\
        \cos\left(\bm B_t t\right)\\
        \sin\left(\bm B_t t\right)
    \end{matrix}\right],
\end{equation}
where $\bm B_x$ and $\bm B_t$ are random projection matrices for spatial and temporal coordinates, respectively. These matrices are independently sampled from Gaussian distributions with hyperparameters $\sigma_x$ and $\sigma_t$. The embedded coordinates $\mathcal{F}(\bm{x}, t)$ are directly fed into the neural network for training and inference.

\subsection{Staggered training scheme for coupled PDE systems}

Multi-physics problems often involve coupled PDE systems with multiple interdependent solution fields, e.g., the phase field corrosion and fracture problems considered in this work. By weighting and summing up all loss terms into a single scalar loss function (see Section \ref{sec:gn-loss-weighting}), the standard training scheme geenrally combines all loss terms associated with different solution fields in a holistic yet undifferentiated manner. However, this standard scheme fails to account for the distinct characteristics and numerical requirements of different solution fields, which may lead to gradient conflicts and suboptimal convergence behavior during training \cite{yuGradientSurgeryMultiTask2020,wangGradientAlignmentPhysicsinformed2025a}. 

\begin{figure}[h]
    \centering
    \includegraphics[width=1.0\textwidth]{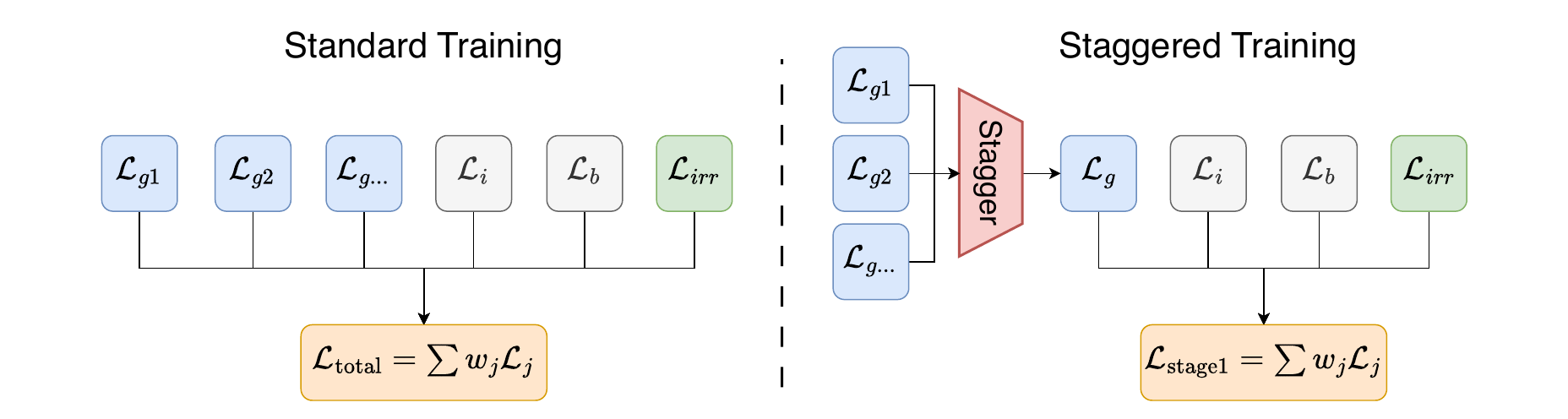}
    \caption{Comparative schematic of standard and staggered training schemes for coupled PDE systems $g_1, g+2, \ldots$. Left: In the standard training scheme, all loss terms are combined into a single scalar loss function and optimized simultaneously. Right: In the staggered training scheme, the loss terms associated with each governing equation are optimized alternately in separate stages. The subscripts $g_1, g_2, b, i, irr$ denote the governing equation, boundary condition, initial condition, and irreversibility regularization losses, respectively. $j$ is the summation index for each loss term.}
    \label{fig:stagger}
\end{figure}

To address this issue, we adopt a staggered training scheme in baseline PINN implementation that alternately optimizes the loss terms associated with each solution field separately \cite{chenSharpPINNsStaggeredHardconstrained2025}. This approach allows the neural network to focus on one solution field at a time, thereby reducing gradient conflicts and improving convergence. Specifically, for a coupled PDE system with several governing equations $g_1, g_2, \ldots$, the training process is divided into multiple stages, where in each stage, only the loss terms related to one governing equation are optimized while simply omitting the other loss terms. Each stage consists of several training iterations, and the stages are cycled through until convergence is achieved for all solution fields. A schematic comparison between the standard and staggered training schemes is illustrated in Figure \ref{fig:stagger}.

It is worth noting that for the steady combustion---an eigenvalue problem---although there is only one governing equation, we still employ the staggered training scheme. Instead of alternating between different governing equations, we alternate between updating the neural network parameters and the eigenvalue parameter. The first stage focuses on optimizing the neural network parameters while keeping the eigenvalue fixed, and the second stage updates the eigenvalue based on the fixed neural network parameters.

\subsection{Causal training strategy for time-dependent problems}

Causality is a fundamental principle in physical systems, dictating that the present state of a system is influenced only by its past states and not by future states. However, standard PINN regardlessly samples collocation points across the entire spatiotemporal domain and computes the point-wise PDE residuals uniformly, without considering the temporal order of information propagation. This indiscriminate treatment can lead to non-causal learning behavior, where the model inadvertently incorporates information from future states into its predictions for the present state, thereby violating the causality principle \cite{wangRespectingCausalityTraining2024}.

Therefore, we incorporate a causal training strategy into the baseline PINN implementation for time-dependent problems. This strategy re-formulates the loss terms associated with the governing equations to respect the temporal order of information propagation. Specifically, at each training iteration, we partition the temporal domain into several sequential time segments and compute the segment-wise MSE of PDE residuals. Then, we weight these segment-wise residuals based on the accumulated residuals from previous segments, thereby prioritizing the learning of earlier time segments before progressing to later ones. The weight calculation for the $i$-th time segment is given by:
\begin{equation}
    w_\text{causal}^i = \exp\left(
        -\epsilon_c \sum_{j=0}^{i-1} \mathcal{L}_g^j(\bm{\theta})
    \right), \text{ for } i = 0, 1, \ldots, N_t
\end{equation}
where $\mathcal{L}_g^j(\bm{\theta})$ is the MSE of PDE residuals for the $j$-th time segment and $N_t$ is the total number of time segments. $\epsilon_c$ is a hyperparameter that controls the strength of causality enforcement. We update $\epsilon_c$ when $w_\text{causal}^{N_t}$ exceeds a predefined threshold (generally set to $0.99$) by multiplying it with a scaling factor (e.g., $2.0$) to progressively enhance the causality effect during training.

\subsection{Gradient-normalized loss weighting}
\label{sec:gn-loss-weighting}

The loss function is formulated as a weighted sum of multiple terms. To balance their respective contributions, we employ a gradient-based weighting strategy \cite{wangExpertsGuideTraining2023,chenGradNormGradientNormalization2018} that dynamically adjusts weights during training. Specifically, at the $s$-th training step, the weight for each loss term is calculated as:
\begin{subequations}
    \begin{gather}
        \hat w_j^{(s)} = \frac{\displaystyle\sum_{j\in\mathcal{J}} \| \nabla_{\bm\theta}\mathcal{L}_j^{(s)} \|}{\| \nabla_{\bm\theta}\mathcal{L}_j^{(s)} \|},
        \quad s \geqslant 1, \quad \forall j\in\mathcal{J}, \\
        w_j^{(s)} = \alpha_w \cdot \hat w_j^{(s-1)} + (1-\alpha_w) \cdot w_j^{(s)} \\
        w_j^{(0)} = 1,
    \end{gather}%
\end{subequations}
where $\mathcal{J}$ denotes the set of all loss terms, typically encompassing PDE residuals, initial/boundary conditions, and in this study, the irreversibility regularization term. $\alpha_w\in[0,1)$ is a smoothing parameter, and $\|\cdot\|$ represents the $L^2$ norm.

\begin{figure}[ht]
    \centering
    \includegraphics[width=0.85\textwidth]{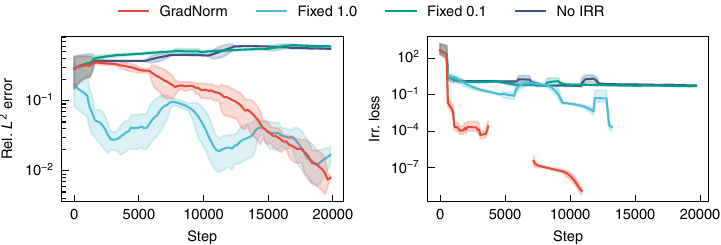}
    \caption{Training histories of relative $L^2$ error of temperature $T$ (left) and irreversibility loss (right) for the steady combustion problem under different weighting configurations of the irreversibility regularization term.}
    \label{fig:combustion-weighting-log}
\end{figure}

The effectiveness of this weighting strategy for balancing PDE residuals and initial/boundary conditions has been extensively validated in previous studies \cite{wangExpertsGuideTraining2023,chenGradNormGradientNormalization2018}. In this work, we extend the same strategy to incorporate the irreversibility regularization term. For the verification purpose, we perform experiments on the combustion problem using four different weighting configurations on the irreversibility term, while keeping all other loss terms consistently weighted by the gradient-based strategy: 1) gradient-based weighting strategy as described above; 2) fixed weight $w_{\text{irr}}=1.0$ for the irreversibility term; 3) fixed weight $w_{\text{irr}}=0.1$ for the irreversibility term; and 4) no irreversibility regularization term (i.e., fixed weight $w_{\text{irr}}=0$). The training histories of irreversibility loss and relative $L^2$ error of temperature $T$ for these configurations are presented in Figure \ref{fig:combustion-weighting-log}, respectively. It can be observed that both the $L^2$ error and irreversibility loss with fixed weights of $0.1$ and $0.0$ plateau at relatively high values. In contrast, the gradient-based weighting strategy and fixed weight of $1.0$ yield significantly lower errors and irreversibility losses, with the gradient-based approach providing faster convergence and a more stable, consistent reduction in $L^2$ error. These results confirm that our proposed irreversibility regularization term can be seamlessly integrated with existing gradient-based weighting strategies.

\subsection{Sampling of collocation points}

For PDE residuals, we employ a random sampling strategy with adaptive refinement \cite{wuComprehensiveStudyNonadaptive2023,wight2020solving}, using two sets of collocation points: 1) a randomly sampled set which is regenerated at regular intervals to ensure global domain coverage; 2) an adaptively refined set updated more frequently based on current residual distributions to target high-error regions. The adaptive refinement $\mathcal{S}_{\text{adapt}}$ is performed by selecting points from a larger pool $\mathcal{S}_{\text{pool}}$ of randomly sampled candidates according to the following criterion:
\begin{equation}
    \mathcal{S}_{\text{adapt}} = \underset{\mathcal{S}\subseteq\mathcal{S}_{\text{pool}}, |\mathcal{S}|=N_{\text{adapt}}}{\arg\max} \sum_{(\bm x,t) \in \mathcal{S}} \left| \mathcal{G}(\bm x, t, \hat u) \right|.
\end{equation}
Here, $N_{\text{adapt}}$ is the number of adaptively refined points to be selected. The adaptive refinement process is repeated at the same frequency as the regeneration of the random set to maintain a balance between exploration and exploitation during training. Unless otherwise specified, we use identical collocation points for both PDE residuals and irreversibility regularization terms.

\subsection{Hyperparameter configurations}
Table \ref{tab:hyperparameters-all} summarizes the hyperparameter configurations for all benchmark problems considered in this study. All hyperparameters are maintained consistently between the baseline PINN and the proposed IRR-PINN implementations to ensure a fair comparison.

\begin{table}[ht]
    \centering
    \caption{Hyperparameter configurations for all benchmark tests.}
    \label{tab:hyperparameters-all}
    \begin{tblr}{
        width=\textwidth,
        colspec={X[0.8,c]X[1.5,c]X[1,c]X[1,c]X[1,c]X[1,c]X[1,c]},
        row{1}={font=\bfseries},
        font=\small
    }
        \hline
        Group & Hyperparameter & TravelingWave & Combustion & IceMelting & Corrosion & Fracture \\
        \hline
        \SetCell[r=6]{c} \textbf{Network}
        & Architecture & ModifiedMLP & ResNet & MLP & ModifiedMLP & ModifiedMLP \\
        & Depth & $6$ & $8$ & $3$ & $6$ & $6$ \\
        & Width & $100$ & $32$ & $64$ & $128$ & $128$ \\
        & Activation & Snake & Tanh & Tanh & GeLU & Swish \\
        & FFE width & $128$ & $64$ & $64$ & $64$ & $0$ \\
        & FFE scale & $2.0\times 2.0$ & $4.0$ & $2.0\times 0.2$ & $1.5\times 1.0$ & --- \\
        \hline
        \SetCell[r=7]{c} \textbf{Training}
        & Epochs & $4000$ & $20000$ & $4000$ & $2500$ & $7000$ \\
        & General points & $20^2$ & $500$ & $20^4$ & $15^3$ & $15^3$ \\
        & Adaptive points & $2000$ & $0$ & $8000$ & $2000$ & $500$ \\
        & Initial LR & $1.0\times 10^{-3}$ & $1.0\times 10^{-3}$ & $5.0\times 10^{-4}$ & $5.0\times 10^{-4}$ & $5.0\times 10^{-4}$ \\
        & Optimizer & Adam & RProp+Adam & Adam & Adam & Adam \\
        & Causality & False & False & True & True & False \\
        & Staggering & False & True & False & True & True \\
        \hline
    \end{tblr}
\end{table}

\begin{figure}[htbp]
    \centering
    \includegraphics[width=0.85\textwidth]{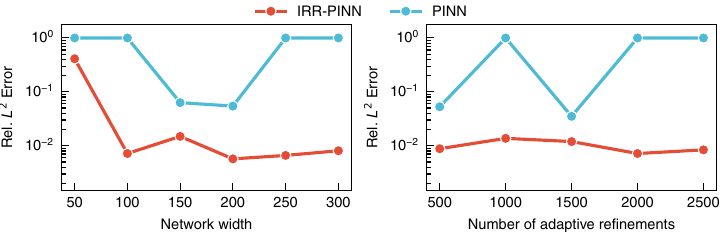}
    \caption{Relative $L^2$ error of traveling wave propagation problem with different network widths (left) and different numbers of adaptive collocation points (right).}
    \label{fig:travellingwave-hyperparameter-sensitivity}
\end{figure}

We also conducted hyperparameter sensitivity analyses for network width and adaptive collocation points on the traveling wave propagation problem. The results, presented in Figure~\ref{fig:travellingwave-hyperparameter-sensitivity}, show that the proposed IRR-PINN consistently outperforms the baseline PINN across different hyperparameter settings and exhibits substantially reduced sensitivity to these choices, demonstrating its robustness and effectiveness.

\section{Detailed description of benchmark problems}

\subsection{Traveling wave propagation}

The traveling wave propagation problem is governed by Fisher-type reaction-diffusion equation. The governing equation is given by:
\begin{equation}
    \frac{\partial u}{\partial t} = D \frac{\partial^2 u}{\partial x^2} + r\,u(1 - \alpha u),
    \label{eq:Fisher}
\end{equation}
subject to the following initial and boundary conditions:
\begin{equation}
    u(x, 0) = A\,\exp\!\left[-\beta(x - x_0)^2\right] \quad \text{and} \quad u(-L/2, t) = u(L/2, t) = 0.
\end{equation}
where $D$ is the diffusion coefficient, $r$ is the reaction rate, $\alpha$ controls the nonlinearity of the reaction term, $A$ and $\beta$ define the amplitude and width of the initial Gaussian profile, and $L$ defines the domain length. For simplicity, the parameters are chosen as $D=1$ $\mathrm{m^2/s}$, $r=1$ $\mathrm{s}^{-1}$, $\alpha=A=1$, $\beta=1$ $\mathrm{m}^{-2}$, $L=40$ m, $x_0=0$, and $T=[0, 20]$ $\mathrm{s}$.

The wave fronts have inherent directional irreversibility as they propagate outward from the initial perturbation, exhibiting an absolute directionality that prevents any backward propagation once the wave has advanced. For the Gaussian initial profile, two fronts will initiate symmetrically from $x_0=0$ and propagate in the $\pm x$ directions. Accordingly, the solution $u(x, t)$ must satisfy opposite irreversibility constraints on either side of the domain, which can be incorporated into a single spatial irreversibility regularization term:
\begin{equation}
    \mathcal{L}_{\text{irr}}^x(\bm{\theta}) = \frac{1}{N_{\text{irr}}} \sum_{j=1}^{N_{\text{irr}}} \text{ReLU}\left( \frac{x_{\text{irr}}^j}{\left\lvert x_{\text{irr}}^j\right\lvert + \epsilon_x} \cdot \frac{\partial \hat{u}}{\partial x}(x_{\text{irr}}^j, t_{\text{irr}}^j; \bm{\theta})\right),
\end{equation}%
with $\epsilon_x>0$ being a small constant for numerical stability.

\subsection{Steady combustion}

Steady combustion is a typical example of directional irreversibility governed by an ordinary differential equation (ODE) system. Here, we consider a freely propagating premixed (FPP) flame in a one-dimensional domain, where the flame front propagates in a single direction. The one-step irreversible chemical reaction is assumed as:
\begin{equation}
    \text{Fuel} + \text{Oxidizer} \rightarrow \text{Products}.\label{eq:combustion-reaction}
\end{equation}%

A simplified set of governing equations for this problem is given by \cite{wuFlamePINN1DPhysicsinformedNeural2025}:
\begin{equation}
    \rho_\text{in}s_\text{L}c_p\frac{\mathrm{d}T}{\mathrm{d}x} - \lambda \frac{\mathrm{d}^2 T}{\mathrm{d}x^2} = -\omega q_F,
\end{equation}
with the supplementary relations:
\begin{subequations}
    \begin{gather}
        \omega = Ae^{-\frac{E_a}{RT}}\left(\rho Y_F\right)^\nu, \\
        u = \frac{c - \sqrt{c^2-4RT/W}}{2}, \\
        c = s_\text{L} + \frac{RT_\text{in}}{Ws_\text{L}}, \\
        \rho = \frac{\rho_\text{in}s_\text{L}}{u}, \\
        Y_F = Y_{F,\text{in}} + \frac{c_p\left(T_\text{in}-T\right)}{q_F},
    \end{gather}
\end{subequations}
where temperature $T(x)$ is the only unknown field to be solved, while other variables (gas density $\rho$, flow velocity $u$, flow pressure $p$, fuel mass fraction $Y_F$, and reaction rate $\omega$) are functions of $T$ and can be derived accordingly. The inlet flow velocity $s_\text{L}$ serves as an eigenvalue of the problem to be determined during the solution process. All other physical constants and parameters are listed in Table~\ref{tab:parameters-combustion}. 

We apply both Dirichlet and Neumann boundary conditions at the inlet ($x=0$) to specify the inflow temperature $T_\text{in}=298\;\mathrm{K}$ and the temperature gradient $\left(\mathrm{d}T / \mathrm{d}x\right)_\text{in}=1.0\times 10^5\;\mathrm{K/m}$, respectively. The domain length is $L=1.5\times 10^{-3}\;\mathrm{m}$.  As the flame front propagates in the positive $x$ direction, the temperature field $T(x)$ must satisfy the forward irreversibility constraint along the spatial dimension. Thus, the irreversibility regularization term can be formulated as:
\begin{equation}
    \mathcal{L}_{\text{irr}}^x(\bm{\theta}) = \frac{1}{N_{\text{irr}}} \sum_{j=1}^{N_{\text{irr}}} \text{ReLU}\left(-\frac{\partial \hat{T}}{\partial x}(x_{\text{irr}}^j; \bm{\theta})\right).
    \label{eq:combustion_irreversibility}
\end{equation}%

\subsection{Ice melting}

We model the ice melting process using a phase field approach based on the Allen--Cahn equation. An idealized, constant temperature field $T\equiv 1$ is prescribed to drive the phase transition, such that the governing equation can be expressed as \cite{jianwangPhaseFieldModelingNumerical2021}: 
\begin{equation}
    \frac{\partial \phi}{\partial t} = M\left(
        \Delta \phi - \frac{F'(\phi)}{\ell^2}  
    \right)- \lambda\frac{\sqrt{2F(\phi)}}{\ell}.
    \label{eq:ice-melting-pde-si}
\end{equation}
where $F(\phi) = \dfrac{1}{4}(\phi^2-1)^2$ is the double-well potential, $M$ is a constant mobility, $\ell$ is the interface thickness, and $\lambda$ is a melting rate parameter. Values of these physical parameters are listed in Table~\ref{tab:parameters-ice-melting}.

We model a three-dimensional case in a cubic domain $\Omega = [-50, 50]^3\;\mathrm{mm}$ over the temporal domain $T = [0, 5]\;\mathrm{s}$. The initial phase field distribution is given by:
\begin{equation}
    \phi(x, y, z, 0) = \tanh\left(\frac{R_0 - \sqrt{x^2 + y^2 + z^2}}{\sqrt{2}\ell}\right),
\end{equation}
where $R_0 = 35\;\mathrm{mm}$ is the initial radius of the ice sphere. Analytically, the melting front advances linearly with time, and the solution for the melting radius is given by:
\begin{equation}
    R(t) = R_0 - \lambda t. \label{eq:ice-melting-analytical}
\end{equation}%

The phase field variable $\phi$ must satisfy the backward irreversibility constraint along the temporal dimension, as ice ($\phi=1$) melts into water ($\phi=-1$), representing a thermodynamically irreversible process where the reverse transition cannot occur spontaneously. To enforce this constraint during PINN training, a temporal irreversibility regularization term is incorporated into the loss function, formulated as:
\begin{equation}
    \mathcal{L}_{\text{irr}}^t(\bm{\theta}) = \frac{1}{N_{\text{irr}}} \sum_{j=1}^{N_{\text{irr}}} \text{ReLU}\left(\frac{\partial \hat{\phi}}{\partial t}(\bm{x}_{\text{irr}}^j, t_{\text{irr}}^j; \bm{\theta})\right).
    \label{eq:melt_irreversibility-si}
\end{equation}

\subsection{Corrosion modeling}
Next, we examine the evolution of pitting corrosion using a KKS (Kim–Kim–Suzuki)–based phase field model \cite{maiPhaseFieldModel2016,cuiPhaseFieldFormulation2021}, in which the metal–electrolyte interface is explicitly represented by a phase-field variable $\phi$ that transitions smoothly from 1 (metal) to 0 (electrolyte) according to the \emph{Allen–Cahn} equation. In addition, a diffusion-type \emph{Cahn–Hilliard} equation is also employed to describe the transport of the normalized metal ion $c$ and to distinguish between activation-controlled and diffusion-controlled corrosion processes. The strong coupling between the \emph{Allen–Cahn} and \emph{Cahn–Hilliard} equations poses a substantial challenge to the accurate prediction of interfacial evolution using PINNs \cite{chenPFPINNsPhysicsinformedNeural2025}, which could in turn highlight the key role of the irreversibility-regularized strategy. The governing equations of this problem are presented directly below, and a more detailed formulation can be found in \cite{chenPFPINNsPhysicsinformedNeural2025,chenSharpPINNsStaggeredHardconstrained2025}.
\begin{subequations}
    \begin{align}
        \text{Cahn--Hilliard: } & \frac{\partial c}{\partial t}- 2\mathcal{A}M \Delta c + 2\mathcal{A}M \left(c_{\mathrm{Se}}-c_{\mathrm{Le}}\right) \Delta h\left(\phi\right) =0 , \label{eq:chcorro-si} \\
        \text{Allen--Cahn: }    & \frac{\partial \phi}{\partial t}
        -2\mathcal{A}L\left[c-h(\phi)\left(c_{\mathrm{Se}}-c_{\mathrm{Le}}\right)-c_{\mathrm{Le}}\right]\left(c_{\mathrm{Se}}-c_{\mathrm{Le}}\right) h^{\prime}(\phi) +L w_\phi g^{\prime}(\phi) - L \alpha_\phi \Delta \phi =0.
        \label{eq:accorro-si}
    \end{align}
    \label{eq:corrosion-pde-simplified-si}
\end{subequations}%
The variables and parameters involved in Equation \eqref{eq:corrosion-pde-simplified-si} are categorized as follows:
\begin{itemize}
    \item Unknown fields: phase field variable $\phi(\bm x, t)$ and normalized concentration $c(\bm x, t)$;
    \item Derived variables: solid and liquid phase concentrations $c_\text{S}(\bm x, t)$ and $c_\text{L}(\bm x, t)$, with $c_\text{S}(\bm x, t)+c_\text{L}(\bm x, t)\equiv1$;
    \item Material constants: $c_\text{Se}$, $c_\text{Le}$, $\mathcal{A}$, $w$, $\alpha_\phi$, $M$, $L$, which are given in Table~\ref{tab:parameters}.
\end{itemize}

The evolution of corrosion is inherently irreversible, as it proceeds unidirectionally from metal ($\phi=1$) to electrolyte ($\phi=0$). Consequently, the phase field variable $\phi$ must satisfy the backward irreversibility constraint along the temporal dimension. Therefore, we apply the same temporal irreversibility regularization term as in Equation \eqref{eq:melt_irreversibility-si} to enforce this constraint during the PINN training.

\begin{figure}[ht]
    \centering
    \includegraphics[width=0.6\textwidth]{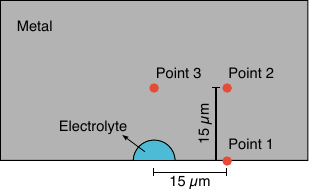}
    \caption{Corrosion modeling. Schematic illustration of the 2D semi-circular pit growth problem. A semi-circular initial pit is defined at the center of the bottom boundary, and the red dots indicate three representative locations where the time evolution of the phase field variable $\phi$ is examined.}
    \label{fig:corrosion-problem-setup}
\end{figure}

We consider a two-dimensional semi-circular pit growth problem, as shown in Figure~\ref{fig:corrosion-problem-setup}. The spatial domain is defined as $\Omega = [-50, 50]\;\mathrm{\mu m} \times [0, 50]\;\mathrm{\mu m}$ and temporal domain as $T = [0, 30]\;\mathrm{s}$. A small initial pit is prescribed at the center of the bottom boundary with $\phi=c=1$. Three monitoring points, marked by red dots in Figure~\ref{fig:corrosion-problem-setup}, are selected to directly examine potential violations with and without the irreversibility constraint in Equation \eqref{eq:melt_irreversibility-si}. 

\subsection{Phase field fracture}
\label{sec:PFF}
Similar to other interfacial phenomena, fracture (cracking) exhibits intrinsic irreversibility in a closed system, evolving monotonically from the intact state ($\phi=0$) to fully damaged ($\phi=1$), as the cracked region cannot recover once initiated. However, unlike other phase field problems, phase field fracture cannot be driven merely by prescribing initial and boundary conditions with different phases; the driving force for crack propagation arises from the strain energy supplied by the external load. Consequently, as the applied stress (or load) increases incrementally, the system undergoes a pronounced, highly nonlinear transition at the crack nucleation threshold, which is very difficult for conventional PINN frameworks to capture. Thus, energy-based approaches, such as \emph{Deep Ritz Method} \cite{manavPhasefieldModelingFracture2024b}, are more commonly adopted within this community. Nevertheless, we believe that our irreversibility-regularized strategy offers a better opportunity to address this challenge using PINNs. 

Phase field fracture involves two strongly coupled fields: the mechanical displacement field $\bm{u}(\bm{x}, t)$, and the phase field variable $\phi(\bm{x}, t)$. The mechanical response of the solid material is described by a degraded linear elastic model, in which the constitutive relationship is modulated by the damage $\phi$. The mechanical equilibrium equation for the displacement field is given by:
\begin{equation}
    \bm\nabla \cdot \left[ g(\phi) \bm \sigma\right] = \bm{0}, \label{eq:mechanical-equilibrium-si}
\end{equation}
and the governing equation for crack evolution reads
\begin{equation}
    r=\frac{G_{c}}{\ell} \left(\phi - \ell^2 \nabla^2 \phi\right)+ g^{\prime}(\phi)\,\psi_{0}(\bm{\varepsilon})\geqslant0,
    \label{eq:fracture-driving-force-si}
\end{equation}
which, together with the irreversibility of the phase field, must satisfy the Karush--Kuhn--Tucker (KKT) conditions:
\begin{equation}
    \dot{\phi}\geqslant0, \quad r\dot{\phi}=0, \quad \text{and}\quad\phi\in[0,1],
\end{equation}
where $g(\phi)=\left(1-\phi\right)^2$ is the degradation function, $\bm \sigma $ is the Cauchy stress tensor, $G_c$ is the critical energy release rate, $\ell$ is the characteristic length scale, and $\psi_{0}(\bm{\varepsilon})$ is the elastic strain energy density. 

The complementarity condition $r\dot\phi=0$ implies that the inequality in Equation \eqref{eq:fracture-driving-force-si} reduces to the equality $r=0$ only on the active set where the damage evolves. In the finite element time-stepping implementation, irreversibility is commonly enforced either by solving a constrained local optimization at each time step (which explicitly satisfies the KKT condition) \cite{Wu2017,Feng2021} or by introducing a history (maximum) driving force to prevent healing \cite{Miehe2010a,Kristensen2021}. Both approaches, however, rely on the time-marching structure and are not directly applicable to an off-line, space–time PINN framework, since the history operator is nonlocal in time and constrained solvers require incremental updates.

To address this limitation, we enforce the KKT conditions in the PINN through a pointwise residual defined as:
\begin{equation}
    \mathcal{R}_\text{KKT}(\bm{x}, t; \bm{\theta}) = 
    \begin{cases}
        \text{ReLU}(-\hat{r}(\bm{x}, t; \bm{\theta})), & \text{if } \left\lvert \dfrac{\partial \hat{\phi}}{\partial t}(\bm{x}, t; \bm{\theta})\right\rvert < \epsilon_{\text{tol}}, \\
        \left\lvert \hat{r}(\bm{x}, t; \bm{\theta}) \right\rvert, & \text{if } \left\lvert \dfrac{\partial \hat{\phi}}{\partial t}(\bm{x}, t; \bm{\theta})\right\rvert \geqslant \epsilon_{\text{tol}} \text{ and } 0 < \hat{\phi}(\bm{x}, t; \bm{\theta}) < 1, \\
        \text{ReLU}(\hat{r}(\bm{x}, t; \bm{\theta})), & \text{if } \hat{\phi}(\bm{x}, t; \bm{\theta}) = 1,
    \end{cases}
    \label{eq:fracture-kkt-residual}
\end{equation}%
where $\epsilon_{\text{tol}}>0$ is a prescribed tolerance, and $\hat{r}(\bm{x}, t; \bm{\theta})$ and $\hat{\phi}(\bm{x}, t; \bm{\theta})$ represent the network predictions of the driving force and damage field, respectively. Additionally, temporal irreversibility is promoted via a regularization term analogous to that used in other phase field problems:
\begin{equation}
    \mathcal{L}_{\text{irr}}^t(\bm{\theta}) = \frac{1}{N_{\text{irr}}} \sum_{j=1}^{N_{\text{irr}}} \text{ReLU}\left(-\frac{\partial \hat{\phi}}{\partial t}(\bm{x}_{\text{irr}}^j, t_{\text{irr}}^j; \bm{\theta})\right).
    \label{eq:fracture_irreversibility}
\end{equation}%

\begin{figure}[ht]
    \centering
    \includegraphics[width=0.55\textwidth]{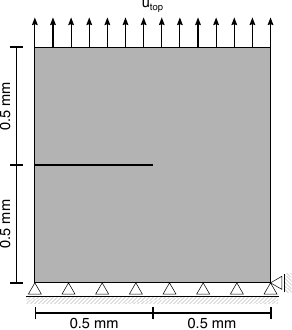}
    \caption{Phase field fracture. Geometric setup of a two-dimensional single-edge notched tension test. An initial crack of length $0.5\;\mathrm{mm}$ is pre-defined from the middle of the left edge. A time-dependent vertical displacement $u_\text{top}(t)$ is applied at the top edge to drive crack propagation.}
    \label{fig:fracture-problem-setup}
\end{figure}

To validate the proposed framework, we consider a paradigmatic benchmark: a two-dimensional single-edge notched tension specimen shown in Figure~\ref{fig:fracture-problem-setup}. Crack initiation and propagation are driven by a prescribed time-dependent vertical displacement $u_\text{top}(t)$ applied to the top boundary. Since our primary focus lies beyond the initial linear elastic response, we adopt a smooth displacement protocol that quickly ramps up and then maintains a constant loading level:
\begin{equation}
    u_\text{top}(t) = u_r \cdot \frac{\tanh\left(\alpha_u t\right)}{\tanh(\alpha_u)}, \quad t \in [0, 1],
\end{equation}
where $u_r$ denotes the target displacement amplitude and $\alpha_u$ governs the transition rate. Material properties for this configuration are detailed in Table~\ref{tab:parameters-fracture}.

Accordingly, the essential initial and boundary conditions are imposed on the top and bottom boundaries through prescribed displacement fields:
\begin{subequations}
    \begin{align}
        u_x(x, y, t) &= \left(y - \frac{H}{2}\right) \left(y + \frac{H}{2}\right) t  \cdot \hat{u}_x, \\
        u_y(x, y, t) &= \left(x - \frac{H}{2}\right) \left(x + \frac{H}{2}\right) t  \cdot \hat{u}_y
        + \frac{1}{H} \left(y - \frac{H}{2}\right) \cdot u_\text{top}(t), 
    \end{align}
\end{subequations}
where $H$ is the specimen height, and $\hat{u}_x$, $\hat{u}_y$ represent the unconstrained displacement predictions of the neural network. The lateral boundaries (left and right edges) are subject to traction-free conditions:
\begin{equation}
    \left(1-\phi\right)^2 \bm{\sigma} \cdot \bm{n} = \bm{0},
\end{equation}
with $\bm{n}$ being the outward unit normal vector. 

For the phase field variable $\phi$, we introduce the following initial condition to represent the pre-defined crack:
\begin{equation}
    \phi(x, y, 0) = 
    \begin{cases}
        \exp\left(\dfrac{-\left\lvert y \right\rvert}{\ell^2}\right) , & \text{if } x \leqslant 0, \\
        0, & \text{if } x > 0.
    \end{cases}
\end{equation}%

\section{Physical parameters}
\label{sec:physical_parameters}

\subsection{Parameters for steady combustion}

\begin{table}[H]
    \centering
    \caption{Physical parameters used in the combustion model (SI units).}
    \label{tab:parameters-combustion}
    \begin{tblr}{width=\textwidth,colspec={X[1,c]X[5,c]X[1,c]},row{1}={font=\bfseries}}
        \hline
        Notation       & Description & Value \\
        \hline
        $R$ & Universal gas constant         & $8.314$               \\
        $A$ & Pre-exponential factor        & $1.4\times 10^8$ \\
        $\nu$ & Reaction order                & $1.6$                   \\
        $E_a$ & Activation energy             & $121417.2$              \\
        $W$ & Molecular weight               & $0.02897$               \\
        $\lambda$ & Thermal conductivity         & $0.026$                 \\
        $c_p$ & Heat capacity        & $1000$                  \\
        $q_F$ & Fuel calorific value         & $5\times 10^7$          \\
        $T_\text{in}$ & Inlet temperature            & $298$           \\
        $(\mathrm{d} T / \mathrm{d} x)_\text{in}$ & Inlet temperature gradient   & $1.0\times 10^{5}$   \\
        $Y_{F,\text{in}}$ & Inlet fuel mass fraction     & $0.0909$\\
        $p_\text{in}$ & Inlet pressure & $101325$ \\
        \hline
    \end{tblr}
\end{table}

\subsection{Parameters for ice melting}

\begin{table}[H]
    \centering
    \caption{Physical parameters used in the ice melting (SI mm units).}
    \label{tab:parameters-ice-melting}
    \begin{tblr}{width=\textwidth,colspec={X[1,c]X[5,c]X[1,c]},row{1}={font=\bfseries}}
        \hline
        Notation       & Description & Value \\
        \hline
        $M$  & Mobility parameter     & $0.1$      \\
        $\ell$ & Interface thickness     & $2.25$      \\
        $\lambda$  & Melting rate parameter  & $5$      \\
        \hline
    \end{tblr}
\end{table}

\subsection{Parameters for corrosion modeling}

\begin{table}[H]
    \centering
    \caption{Physical parameters used in the corrosion modeling (SI units).}
    \label{tab:parameters}
    \begin{tblr}{width=\textwidth,colspec={X[1,c]X[5,c]X[1,c]},row{1}={font=\bfseries}}
        \hline
        Notation       & Description & Value \\
        \hline
        $\alpha_\phi$   & Gradient energy coefficient                               & $1.03\times10^{-4}$  \\
        $w_\phi$        & Height of the double well potential                       & $1.76\times 10^7$    \\
        $\ell$          & Interface thickness                                       & $1.0\times 10^{-5}$   \\
        $L$             & Mobility parameter                                        & $2.0$                \\
        $M$             & Diffusivity parameter                                     & $7.94\times10^{-18}$ \\
        $\mathcal{A}$   & Free energy density-related parameter                     & $5.35\times 10^7$    \\
        $c_\mathrm{Se}$ & Normalised equilibrium concentration for the solid phase  & $1.0$                \\
        $c_\mathrm{Le}$ & Normalised equilibrium concentration for the liquid phase & $0.036$              \\
        \hline
    \end{tblr}
\end{table}

\subsection{Parameters for the phase field fracture}

\begin{table}[H]
    \centering
    \caption{Physical parameters used in the phase field fracture model (SI mm units).}
    \label{tab:parameters-fracture}
    \begin{tblr}{width=\textwidth,colspec={X[1,c]X[5,c]X[2,c]},row{1}={font=\bfseries}}
        \hline
        Notation       & Description & Value \\
        \hline
        $E$     & Young's modulus               & $210\times 10^3$      \\
        $\nu$   & Poisson's ratio               & $0.3$                 \\
        $\lambda$ & Lamé's first parameter       & $121.153\times 10^3$  \\
        $\mu$   & Shear modulus                 & $80.769\times 10^3$   \\
        $G_c$   & Critical energy release rate  & $2.7$                 \\
        $\ell$  & Characteristic length scale   & $0.024$                \\
        $u_r$   & Final displacement magnitude  & $0.0525$               \\
        $\alpha_u$& Ramping speed of displacement function & $4.0$      \\
        \hline
    \end{tblr}
\end{table}

\section{Numerical implementation of reference solutions}

We provide reference solutions for all benchmark problems using conventional numerical methods, which serve as ground truth for evaluating the accuracy of the PINN predictions. Except for the steady combustion problem, which is solved using a shooting method as detailed in \cite{wuFlamePINN1DPhysicsinformedNeural2025}, all other problems are solved using the finite element method (FEM) implemented in the open-source library \texttt{FEniCS} \cite{alnaes2015fenics}. 

\subsection{Traveling wave propagation}

We introduce a test function $v$ and apply backward Euler time discretization to derive the weak form of Equation \eqref{eq:Fisher} as:
\begin{equation}
    \int_{\Omega} \frac{u^{n+1}-u^{n}}{\Delta t} v \,\mathrm{d}\Omega
    + D\int_{\Omega} \nabla u^{n+1} \cdot \nabla v \,\mathrm{d}\Omega
    - r\int_{\Omega} u^{n+1}(1 - \alpha u^{n+1}) v \,\mathrm{d}\Omega = 0,
\end{equation}
where $u^n$ and $u^{n+1}$ denote the solution at the current and next time steps, respectively, and $\Delta t$ is the time step size. We discretize the spatial domain using linear Lagrange elements with a total of 1000 elements. The time step size is set to $\Delta t = 0.02$ s.

\subsection{Steady combustion}
The reference solution for the steady combustion problem is obtained using a shooting method with bisection eigenvalue search \cite{wuFlamePINN1DPhysicsinformedNeural2025}, as outlined in Algorithm~\ref{alg:shooting_method_combustion}. The spatial domain is discretized into 1000 uniform grid points.

\begin{algorithm}[htbp]
\caption{Shooting method with bisection eigenvalue search for steady combustion problem.}
\label{alg:shooting_method_combustion}
\DontPrintSemicolon
\SetKw{KwAnd}{and}
\SetKw{KwOr}{or}
\SetKwInput{KwGiven}{Given}
\SetKwInput{KwRequire}{Require}
Discretize spatial domain: $x_i = i\Delta x,\; i=0,\dots,n-1$, with $\Delta x = \dfrac{L}{n-1}$.\;
Initialize arrays: $T_i,\ (\nabla T)_i,\ u_i,\ \rho_i,\ Y_{F,i},\ \omega_i,\ p_i$.\;
Impose inlet conditions:
\[
T_0 = T_{\text{in}},\quad (\nabla T)_0 = (\nabla T)_{\text{in}},\quad p_0=p_{\text{in}},\quad
Y_{F,0}=Y_{F,\text{in}},\quad \rho_0=\rho_{\text{in}},\quad
\omega_0 = A e^{-E_a/(R T_0)}(\rho_0 Y_{F,0})^\nu.
\]\\
Set initial bisection bracket for eigenvalue $s_L^{(l)}, s_L^{(r)}$.\;
\For{$k=0,1,2,\dots$}{
    Calculate midpoint as eigenvalue estimate: $s_L^{(k)} = (s_L^{(l)} + s_L^{(r)}) / 2$ .\;
    Precompute auxiliary constants:
    \[
    c_1 = \Delta x \frac{\rho_{\text{in}} c_p}{\lambda} s_L^{(k)},\quad
    c_2 = \Delta x \frac{q_F}{\lambda},\quad
    c_3 = s_L^{(k)} + \frac{R_g T_{\text{in}}}{s_L^{(k)}}.
    \]\\
    Set $u_0 = s_L^{(k)}$.\;
    Flag $\mathtt{converged} \gets \text{true}$.\;
    \For{$i=1$ \KwTo $n-1$}{
        Forward update of temperature and temperature gradient:
        \[
        (\nabla T)_i = (\nabla T)_{i-1} + c_1 (\nabla T)_{i-1} - c_2 \omega_{i-1},
        \quad
        T_i = T_{i-1} + \Delta x\, (\nabla T)_i.
        \]
        \If{$(\nabla T)_i < 0$ (flashback)}{
            $s_L^{(l)} \gets s_L^{(k)}$\;
            $\mathtt{converged}\gets \text{false}$\;
            \textbf{break}\;
        }
        \ElseIf{$T_i > T_{\max}$ (blow-off)}{
            $s_L^{(r)} \gets s_L^{(k)}$
            $\mathtt{converged}\gets \text{false}$\;
            \textbf{break}\;
        }
        \Else{
            Update flow variables:
            \[
            u_i = \frac{c_3 - \sqrt{c_3^2 - 4 R_g T_i}}{2},
            \quad
            \rho_i = \rho_{\text{in}} \frac{s_L^{(k)}}{u_i},
            \quad
            p_i = \rho_i R_g T_i,
            Y_{F,i} = Y_{F,\text{in}} + \frac{c_p (T_{\text{in}} - T_i)}{q_F},
            \quad
            \omega_i = A e^\frac{-E_a}{R T_i}(\rho_i Y_{F,i})^{\nu}.
            \]\\
        }
    }
    \If{$\mathtt{converged} = \text{true}$ \KwOr $|s_L^{(r)} - s_L^{(l)}| < \varepsilon$}{
        \If{$i < n-1$}{
            Fill remaining indices $j=i,\dots,n-1$ with equilibrium tail:
            \[
            T_j \gets T_{i-1},\quad (\nabla T)_j \gets 0,\quad
            (u_j,\rho_j,p_j,Y_{F,j},\omega_j) \gets (u_{i-1},\rho_{i-1},p_{i-1},Y_{F,i-1},\omega_{i-1}).
            \]
        }
        \textbf{break}
    }
}
\BlankLine
\textbf{Output:} Eigenvalue estimate $s_L^\star \approx s_L^{(k)}$, profiles $T(x), Y_F(x), u(x), \rho(x), p(x), \omega(x), (\nabla T)(x)$.
\end{algorithm}

\subsection{Ice melting}

The weak form of the Allen--Cahn equation (Equation \eqref{eq:ice-melting-pde-si}) is derived as:
\begin{equation}
    \int_{\Omega} \frac{\phi^{n+1}-\phi^{n}}{\Delta t} v \,\mathrm{d}\Omega
    + M \int_{\Omega} \nabla \phi^{n+1} \cdot \nabla v \,\mathrm{d}\Omega
    + \frac{M}{\ell^2} \int_{\Omega} F'(\phi^{n+1}) v \,\mathrm{d}\Omega
    + \frac{M\lambda}{\ell} \int_{\Omega} \sqrt{2F(\phi^{n+1})} v \,\mathrm{d}\Omega = 0,
\end{equation}
We discretize the cubic spatial domain using linear Lagrange elements with a total of $64^3$ elements and time step size $\Delta t = 0.005$ s.

\subsection{Corrosion modeling}
The weak forms of the Cahn--Hilliard and Allen--Cahn equations (Equations \eqref{eq:chcorro-si} and \eqref{eq:accorro-si}) with backward Euler time discretization and test functions $v_c$ and $v_\phi$ are given by:
\begin{equation}
    \int_{\Omega} \frac{c^{n+1}-c^{n}}{\Delta t} v_c \,\mathrm{d}\Omega
    + 2\mathcal{A}M \int_{\Omega} \nabla c^{n+1} \cdot \nabla v_c \,\mathrm{d}\Omega
    - 2\mathcal{A}M \int_{\Omega} \left(c_{\mathrm{Se}}-c_{\mathrm{Le}}\right) \nabla h\left(\phi^{n+1}\right) \cdot \nabla v_c \,\mathrm{d}\Omega = 0,
    \label{eq:cahn-hilliard-weak-form}
\end{equation}
and
\begin{equation}
    \begin{aligned}
    \int_{\Omega} \frac{\phi^{n+1}-\phi^{n}}{\Delta t} v_\phi \,\mathrm{d}\Omega
    - 2\mathcal{A}L \int_{\Omega} \left[c^{n+1}-h(\phi^{n+1})\left(c_{\mathrm{Se}}-c_{\mathrm{Le}}\right)-c_{\mathrm{Le}}\right]\left(c_{\mathrm{Se}}-c_{\mathrm{Le}}\right) h^{\prime}(\phi^{n+1}) v_\phi \,\mathrm{d}\Omega
    &\\
    + L w_\phi \int_{\Omega} g^{\prime}(\phi^{n+1}) v_\phi \,\mathrm{d}\Omega
    - L \alpha_\phi \int_{\Omega} \nabla \phi^{n+1} \cdot \nabla v_\phi \,\mathrm{d}\Omega &= 0.    
    \label{eq:allen-cahn-weak-form}
    \end{aligned}
\end{equation}
We directly sum Equations \eqref{eq:cahn-hilliard-weak-form} and \eqref{eq:allen-cahn-weak-form} to form a coupled nonlinear system, which is solved using the Newton-Raphson method at each time step. We discretize the spatial domain using linear Lagrange elements with a total of $100 \times 50$ elements. An adaptive time-stepping scheme is employed with an initial time step size of $\Delta t = 0.001$ s. Algorithm \ref{alg:fenics} provides a general outline of the \texttt{FEniCS} implementation for the phase field model of pitting corrosion.
\begin{algorithm}[htbp]
    \caption{\texttt{FEniCS} implementation for the phase field model of pitting corrosion}
    \label{alg:fenics}
    Initialize mesh, function spaces, and model parameters\;
    Define the boundary conditions and initial conditions\;
    Formulate the weak forms of governing equations according to Eqs. \eqref{eq:cahn-hilliard-weak-form} and \eqref{eq:allen-cahn-weak-form}\;

    Initialize time $t = 0$ and time step $\Delta t$\;
    \While{
        $t < T_{\text{final}}$
    }{
        Try solving the coupled nonlinear system\;
        \eIf{
            Converged
        }{
            Update $c$ and $\phi$ according to the solution\;
            Step forward $t \leftarrow t + \Delta t$\;
            \If {
                $n_\text{iter} < 6$
            }{
                Increase time step $\Delta t \leftarrow 2 \Delta t$\;
            }
        }{
            Decrease time step $\Delta t \leftarrow \Delta t / 2$\;
            Retry solving\;
            \If {
                $\Delta t < \Delta t_{\text{min}}$
            }{
                Break\;
            }
        }
    }
\end{algorithm}%
where $T_{\text{final}}$ is the final simulation time, $n_\text{iter}$ is the number of iterations taken to converge at the current time step, and $\Delta t_{\text{min}}$ is the minimum allowable time step size.

\subsection{Phase field fracture}

We employ a staggered scheme to solve the coupled system of Equations \eqref{eq:fracture-mechanical-equilibrium} and \eqref{eq:fracture-driving-force-si}. At each time step, we first solve the mechanical equilibrium equation for the displacement field $\bm{u}$ with a fixed phase field $\phi$, followed by solving the phase field evolution equation for $\phi$ with the updated displacement field $\bm{u}$. Each subproblem is solved using the linear solver in \texttt{FEniCS}. The weak forms of the mechanical equilibrium equation with test function $\bm{v}_u$ is given by:
\begin{equation}
    \int_{\Omega} g(\phi^{n}) \bm{\sigma}(\bm{u}^{n+1}) : \bm{\varepsilon}(\bm{v}_u) \,\mathrm{d}\Omega = 0,
\end{equation}
To solve the phase field evolution equation, we introduce a history field $H^+$ to enforce the numical irreversibility, defined as:
\begin{equation}
    H^+ = \max_{t\in[0, \tau]}\psi_0^+(\bm{\varepsilon}(\bm{u}(t))),
\end{equation}
where $\psi_0^+(\bm{\varepsilon})$ is the tensile part of the elastic strain energy density \cite{Amor2009}, defined as:
\begin{equation}
    \psi_0^+(\bm{\varepsilon}) = \frac{1}{2} K \langle \text{tr}(\bm{\varepsilon}) \rangle_+^2 + \mu \left(
        \bm{\varepsilon}^\text{dev} : \bm{\varepsilon}^\text{dev}
    \right)
\end{equation}
with $\langle \cdot \rangle_+ = \max(0, \cdot)$ being the positive part operator, $K$ the bulk modulus, and $\bm{\varepsilon}^\text{dev}$ the deviatoric strain tensor. The weak form of the phase field evolution equation with test function $v_\phi$ is given by:
\begin{equation}
    G_c \ell \int_{\Omega} \nabla \phi^{n+1} \cdot \nabla v_\phi \,\mathrm{d}\Omega
    + \frac{G_c}{l} \int_{\Omega} \phi^{n+1} v_\phi \,\mathrm{d}\Omega
    - \int_{\Omega} 2(1 - \phi^{n+1}) H^+ v_\phi \,\mathrm{d}\Omega = 0.
\end{equation}

\begin{figure}[ht]
    \centering
    \includegraphics[width=0.6\textwidth]{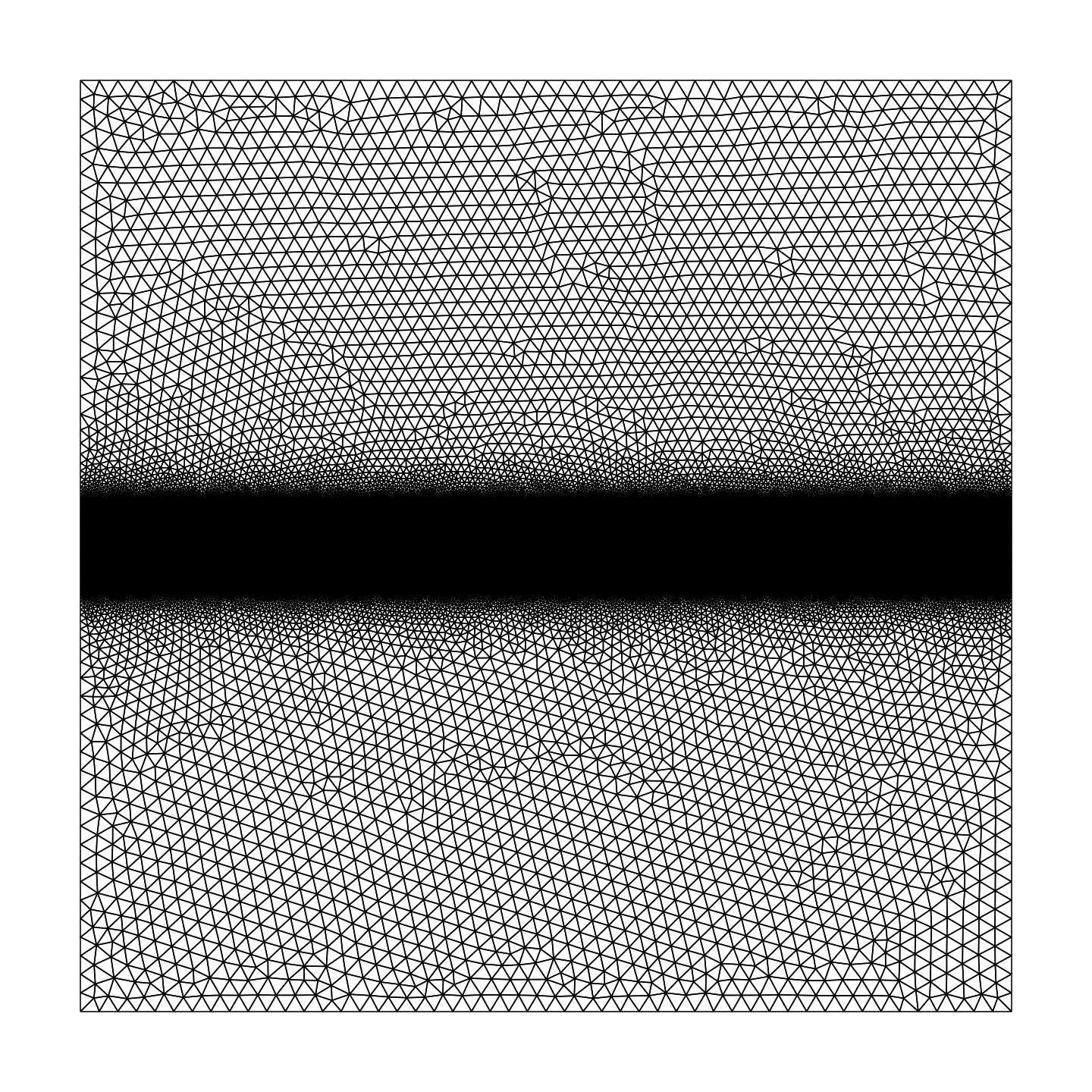}
    \caption{Phase field fracture. Finite element mesh for the single-edge notched tension test. Local mesh refinement is applied around the notch tip to accurately capture the stress concentration and crack propagation.}
    \label{fig:fracture-mesh}
\end{figure}

To discretize the spatial domain, we employ the \texttt{Gmsh} software to generate an unstructured triangular mesh with a local refinement around the notch tip. The global corner size is set to 0.02 mm and the local mesh size is refined to 0.002 mm near the notch tip, resulting in a total of 77,050 elements, as illustrated in Figure~\ref{fig:fracture-mesh}. The loading step size is set to $3.5\times 10^{-4}\,\mathrm{mm}$ in the elastic regime and reduced to $7.0\times 10^{-7} \,\mathrm{mm}$ after crack initiation to accurately capture the crack propagation process.

\section{Computational cost}
\label{sec:computational_cost}

The overall PINN framework for all benchmark problems is implemented in \texttt{Python} using the \texttt{JAX} ecosystem, specifically leveraging \texttt{Flax} for neural network architectures and \texttt{Optax} for gradient-based optimization with automatic differentiation capabilities. Reference solutions for the steady combustion problem are computed using \texttt{Python}, while all other reference solutions are obtained using \texttt{FEniCS} implemented in \texttt{C++} with \texttt{Python} bindings. All numerical experiments are conducted on a high-performance computing platform equipped with an AMD EPYC 7543 32-core processor with 80 GB RAM and an NVIDIA A40 GPU with 48 GB VRAM, ensuring computational efficiency for both PINN training and FEM reference calculations. Table~\ref{tab:computational_cost} summarizes the computational costs for training the PINN models and obtaining reference solutions for each benchmark problem. We observe that the irreversibility regularization term introduces only a marginal increase in training time compared to baseline PINN models. Considering the significant performance improvements achieved through irreversibility regularization, this additional computational overhead is well justified. Notably, while PINN models require longer training times than reference solutions for smaller-scale problems such as traveling wave propagation and steady combustion (a common issue of PINN applications), for problems involving large element counts and extended simulation times, such as ice melting and phase field fracture, PINN models demonstrate substantial computational advantages over traditional numerical methods in obtaining accurate solutions.

\begin{table}[ht]
    \centering
    \caption{Computational cost for training PINN models and obtaining reference solutions for all benchmark problems (s).}
    \label{tab:computational_cost}
    \begin{tblr}{
        width=\textwidth,
        colspec={X[1.5,c]X[1,c]X[1,c]X[1,c]X[1,c]X[1,c]},
        row{1}={font=\bfseries},
        font=\small
    }
        \hline
        Model & TravelingWave & Combustion & IceMelting & Corrosion & Fracture \\
        \hline
        IRR-PINN & 391 & 632 & 2276 & 1337 & 529 \\
        Baseline PINN & 385 & 617 & 2083 & 1329 & 536 \\
        Reference Solution & 2.40 & 1.50 & 65175 & 91 & 2960 \\
        \hline
    \end{tblr}
\end{table}

\putbib[bibfile-cnx,bibfile-ccj,bibfile-irr,bibfile] 
\end{bibunit}
\label{SIEnd}
\end{document}